\documentclass[10pt,twocolumn,letterpaper]{article}

\usepackage[pagenumbers]{cvpr} %

\usepackage[dvipsnames]{xcolor}

\definecolor{darkgreen}{RGB}{0,100,0}
\definecolor{darkred}{RGB}{139,0,0}
\definecolor{darkblue}{RGB}{0,0,139}

\usepackage[utf8]{inputenc} %
\usepackage[T1]{fontenc}    %
\usepackage{amsfonts}       %
\usepackage{nicefrac}       %
\usepackage{microtype}      %

\usepackage{graphicx}
\usepackage{amsmath,amssymb}
\usepackage{amsthm}
\usepackage{multirow}
\usepackage{multicol}
\usepackage{booktabs}
\usepackage{longtable}

\usepackage{url}
\usepackage{courier}
\usepackage{caption}
\usepackage{times}
\usepackage{epsfig}
\usepackage{graphicx}
\usepackage{verbatim}
\usepackage{bbm}

\usepackage{enumitem,kantlipsum}

\usepackage{algorithmic}
\usepackage{lipsum}

\usepackage{wrapfig}
\usepackage{algorithm}
\usepackage{listings}
\usepackage{etoolbox}
\makeatletter
\AfterEndEnvironment{algorithm}{\let\@algcomment\relax}
\AtEndEnvironment{algorithm}{\kern2pt\hrule\relax\vskip3pt\@algcomment}
\let\@algcomment\relax
\newcommand\algcomment[1]{\def\@algcomment{\footnotesize#1}}
\renewcommand\fs@ruled{\def\@fs@cfont{\bfseries}\let\@fs@capt\floatc@ruled
  \def\@fs@pre{\hrule height.8pt depth0pt \kern2pt}%
  \def\@fs@post{}%
  \def\@fs@mid{\kern2pt\hrule\kern2pt}%
  \let\@fs@iftopcapt\iftrue}
\makeatother

\usepackage{tabularx}
\usepackage{rotating}
\usepackage{dashrule}
\usepackage{color, colortbl}

\newlength\savewidth\newcommand\shline{\noalign{\global\savewidth\arrayrulewidth
  \global\arrayrulewidth 1pt}\hline\noalign{\global\arrayrulewidth\savewidth}}

\newcommand{\datatagnew}[1]{\rotatebox[origin=l]{90}{\scriptsize{#1}}}
\newcolumntype{C}{>{\centering\arraybackslash}X}
\definecolor{OurBG}{rgb}{0.9, 0.9, 1.}

\newcolumntype{x}[1]{>{\centering\arraybackslash}p{#1pt}}
\newcolumntype{y}[1]{>{\raggedright\arraybackslash}p{#1pt}}
\newcolumntype{z}[1]{>{\raggedleft\arraybackslash}p{#1pt}}

\usepackage{amsmath,amsfonts,bm}

\def\eqref#1{equation~\ref{#1}}

\def\1{\bm{1}}

\def\rvp{{\mathbf{p}}}
\def\rvq{{\mathbf{q}}}

\DeclareMathAlphabet{\mathsfit}{\encodingdefault}{\sfdefault}{m}{sl}
\SetMathAlphabet{\mathsfit}{bold}{\encodingdefault}{\sfdefault}{bx}{n}

\definecolor{Gray}{gray}{0.5}
\definecolor{nicergreen}{rgb}{0.13, 0.54, 0.13}
\definecolor{nicered}{rgb}{0.83, 0.16, 0.16}
\definecolor{Highlight}{HTML}{39b54a}  %

\newcommand\name{SynCLR}

\definecolor{cvprblue}{rgb}{0.21,0.49,0.74}
\usepackage[pagebackref,breaklinks,colorlinks,citecolor=cvprblue]{hyperref}

\title{Learning Vision from Models Rivals Learning Vision from Data}

\author{
   Yonglong Tian${^{1, \dagger}}$\hspace{2mm} Lijie Fan${^{2, \dagger, }}$\thanks{Work done while interning at Google.}\hspace{2mm} Kaifeng Chen$^1$\hspace{2mm} Dina Katabi$^2$\hspace{2mm} Dilip Krishnan$^1$\hspace{2mm} Phillip Isola$^2$\hspace{2mm}\\[3mm]
  $^1$Google Research, \hspace{3pt}
  $^2$MIT CSAIL, \hspace{3pt}
  $^\dagger$equal contribution\\[2mm]
  {\small \hspace{3pt} Github Repo: 
  \url{https://github.com/google-research/syn-rep-learn}}
}

\begin{document}
\maketitle
\begin{abstract}
We introduce \name, a novel approach for learning visual representations exclusively from synthetic images and synthetic captions, without any real data. We synthesize a large dataset of image captions using LLMs, then use an off-the-shelf text-to-image model to generate multiple images corresponding to each synthetic caption. We perform visual representation learning on these synthetic images via contrastive learning, treating images sharing the same caption as positive pairs. The resulting representations transfer well to many downstream tasks, competing favorably with other general-purpose visual representation learners such as CLIP and DINO v2 in image classification tasks. Furthermore, in dense prediction tasks such as semantic segmentation, \name~outperforms previous self-supervised methods by a significant margin, \eg, improving over MAE and iBOT by 6.2 and 4.3 mIoU on ADE20k for ViT-B/16.

\end{abstract}    
\section{Introduction}
\label{sec:intro}

Representation learning extracts and organizes information from raw, often unlabeled data. The quality, quantity, and diversity of the data determines how good a representation the model can learn. The model becomes a reflection of the collective intelligence that exists in the data. We get what we feed in.

Unsurprisingly, the current best-performing visual representation learning methods~\cite{clip,dinov2} rely on large scale real datasets. However, the collection of real data has its own dilemmas. Collecting \emph{large scale uncurated} data~\cite{laion} is relatively cheap and thus quite achievable. However, for self-supervised representation learning, this approach exhibits poor scaling behavior --i.e., adding more uncurated data has little effect at large data scales~\cite{sslscaling,DnC}. Collecting \emph{small scale curated} data~\cite{imagenet} also is achievable, but models trained in this way are limited to relatively narrow tasks. The ideal would be large scale curated datasets of real images, and recent work has indeed shown that this can lead to strong performance gains at scale~\cite{dinov2}, but this path is costly to pursue.

\begin{figure}[t]
\centering
\small
\includegraphics[width=0.98\linewidth]{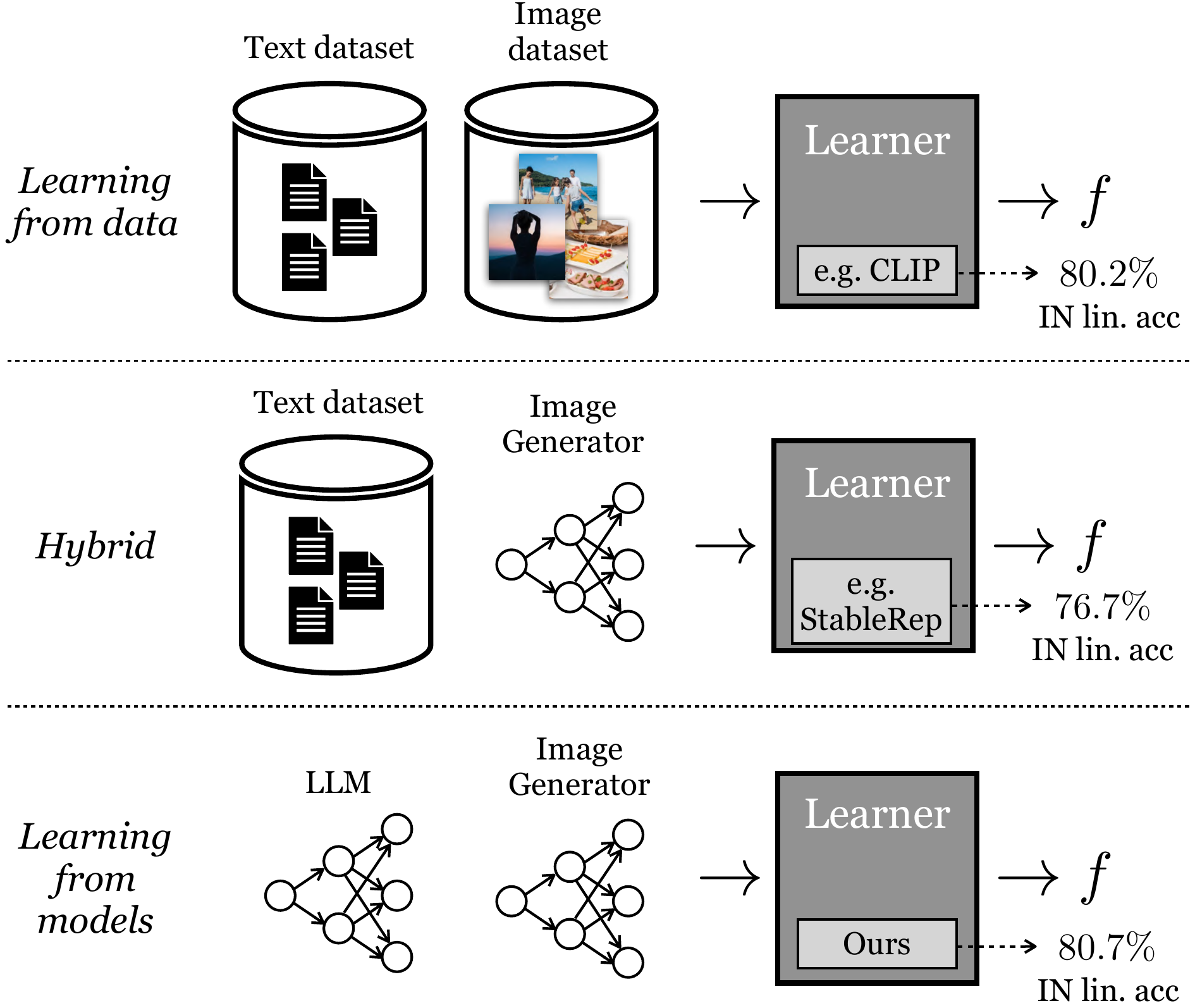}
\caption{Three paradigms for visual representation learning. Top row: Traditional methods, such as CLIP~\cite{clip}, learn only from real data; Middle row: Recent methods, such as  StableRep~\cite{stablerep}, learn from real text and generated images; Bottom row: Our method, SynCLR, learns from synthetic text and synthetic images, and rival the linear transfer performance of CLIP on ImageNet despite not directly observing any real data.}
\label{fig:paradigm}
\vspace{-5pt}
\end{figure}

To alleviate the cost, in this paper we ask if \textit{synthetic data}, sampled from off-the-shelf generative models, is a viable path toward large scale curated datasets that can train state-of-the-art visual representations.

We call such a paradigm \emph{learning from models}, in contrast to directly \emph{learning from data}. Models have several advantages as a data source for building large scale training sets: via their latent variables, conditioning variables, and hyperparameters, they provide new controls for curating data; we will make use of these controls in the method we propose. Models also can be easier to share and store (because models are more compressed than data), and can produce an unlimited number of data samples (albeit with finite diversity). A growing literature has studied these properties and other advantages (and disadvantages) of using generative models as a data source for training downstream models~\cite{jahanian2021generative,he2022synthetic,azizi2023synthetic,fake,stablerep,scaling}. Some of these methods use a \emph{hybrid} mode -- either mixing real and synthetic datasets~\cite{azizi2023synthetic} or needing a real dataset to generate another synthetic dataset~\cite{stablerep}. Other methods try to learn representations from purely synthetic data~\cite{fake} but lag far behind the best performing models. Instead, we show that \emph{learning from models}, without training on any real data,
can yield representations that match the top-performing representations learnt from real data. For instance, as illustrated in Figure~\ref{fig:paradigm}, representations learnt by our method are able to transfer as well as OpenAI's CLIP~\cite{clip} on ImageNet (both methods using ViT-B~\cite{vit}). %

\begin{figure}[t]
\centering
\small
\includegraphics[width=1.0\linewidth]{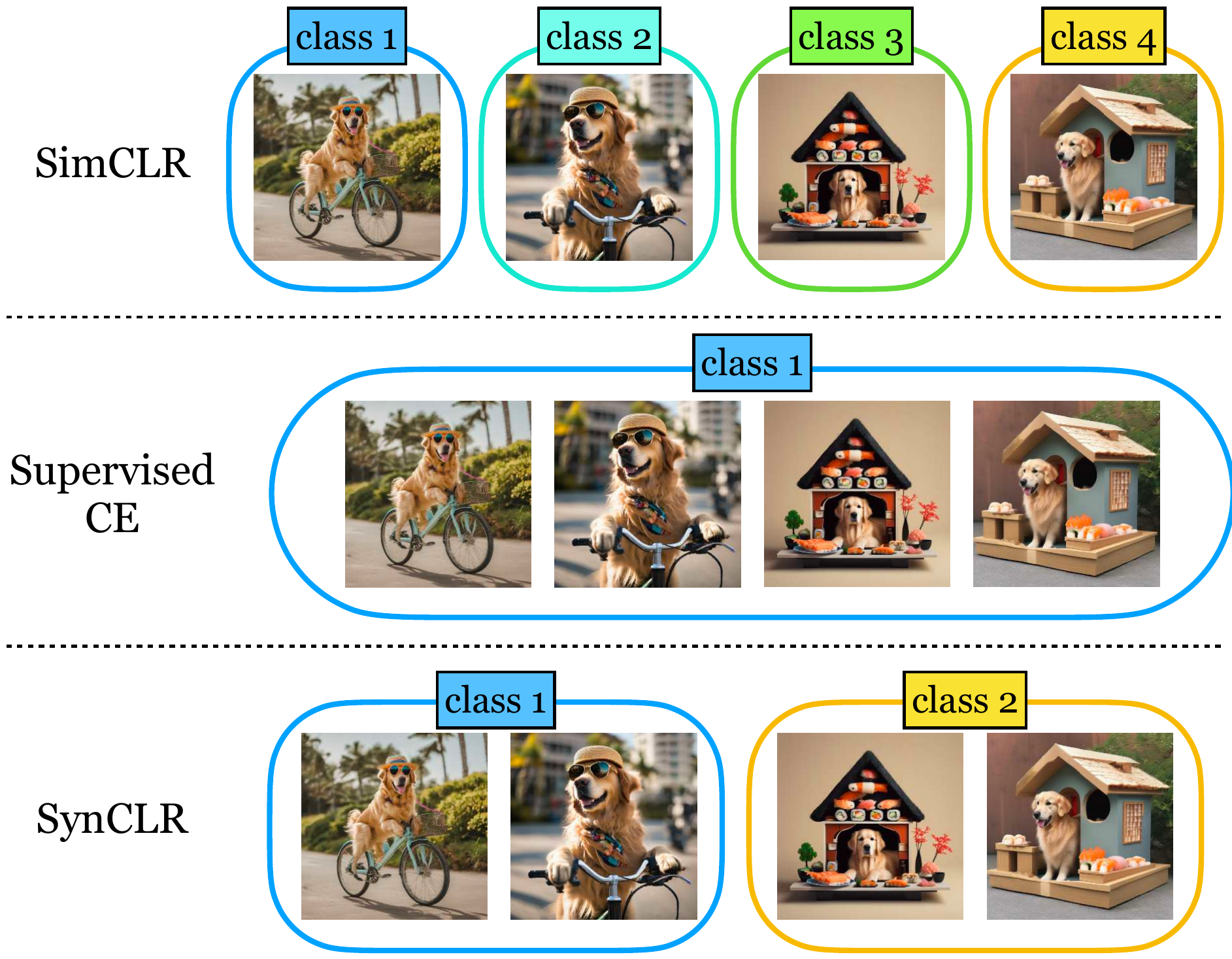}
\caption{Different learning objectives treat classification granularity differently. These images are generated by two prompts ``\emph{a golden retriever, wearing sunglasses and a beach hat, rides a bike}" and ``\emph{a cute golden retriever sits in a house made of sushi}".
 SimCLR treats each image as a class, while supervised cross-entropy treats them all as the same ``golden retrieval'' class. The former does not consider shared semantics between images, and the latter is coarse-grained and ignores actions or relationships between subjects/background. Our approach, SynCLR, defines visual classes by sentences.}
\label{fig:model_diff}
\vspace{-5pt}
\end{figure}

Our approach leverages generative models to re-define the granularity of visual classes. As shown in Figure~\ref{fig:model_diff}, consider we have four images generated using two prompts: ``\emph{a golden retriever, wearing sunglasses and a beach hat, rides a bike}" and ``\emph{a cute golden retriever sits in a house made of sushi}". Traditional self-supervised method such as SimCLR~\cite{simclr} will treat each of these images as a different class;  embeddings for different images are pushed apart with no explicit consideration of the shared semantics between images. On the other extreme, supervised learning methods (\ie SupCE) will regard all these images as a single class (e.g., ``golden retriever''). This ignores nuances in the semantics of the images, such as the fact that the dogs are riding a bike in one pair of images and sitting inside a sushi house in the other pair of images. Instead, our method, \name, treats \textit{captions} as classes, \ie, each caption describes a visual class (this level of granularity was also explored in StableRep~\cite{stablerep}). This allows us to group images by the concepts of ``riding a bike'' and ``sitting in a sushi house'', in addition to grouping by a coarser class label like ``golden retrieval''. This level of granularity is difficult to mine in real data, since collecting multiple images described by a given caption is non-trivial, especially when scaling up the number of captions. However, text-to-image diffusion models are fundamentally built with this ability; simply by conditioning on the same caption and using different noise inputs, a text-to-image diffusion model will produce different images that all match the same caption. In our experiments, we find the caption-level granularity outperforms both SimCLR and supervised training. Another advantage is that this definition of visual classes has good scalability. Unlike ImageNet-1k/21k where a given number of classes is fixed, we can augment existing classes (or data) in an online fashion, and theoretically scale up to as many classes as needed.

Our system consists of three steps. The first step is to synthesize a large corpus of image captions. We design a scalable approach by leveraging the in-context learning capability of large language models (LLMs), where we present examples of word-to-caption translations. Next, a text-to-image diffusion model is adopted to synthesize multiple images for each synthetic caption. This yields a synthetic dataset of 600M images. Then we train visual representation models by a combination of multi-positive contrastive learning~\cite{supcon} and masked image modeling~\cite{ibot}. 

Our learned representations transfer well. With~\name~pre-training, our ViT-B and ViT-L models achieve 80.7$\%$ and 83.0$\%$ top-1 linear probing accuracy on ImageNet-1K, respectively, which is on par with OpenAI's CLIP~\cite{clip}. On fine-grained classification tasks, \name~outperforms CLIP by 3.3$\%$ for ViT-B and 1.5$\%$ for ViT-L, and performs similarly to DINO v2~\cite{dinov2} models, which are distilled from a pre-trained ViT-g model.  For semantic segmentation on ADE20k, ~\name~outperforms MAE pre-trained on ImageNet by 6.2 and 4.1 in mIoU for ViT-B and ViT-L under the same setup, showing  strong transfer ability for dense prediction tasks similar to DINO v2, which additionally involves a training period on 518x518 resolution images that~\name~does not have.

\section{Related Works}

\textbf{Self-supervised representation learning} approaches in vision develop domain-specific pre-text tasks, such as colorization~\cite{zhang2016colorful}, rotation prediction~\cite{gidaris2018unsupervised}, and solving jigsaw puzzles~\cite{noroozi2016unsupervised}. Domain-agnostic approaches have been popular, such as contrastive learning~\cite{becker1992self,hadsell2006dimensionality,cpc,insdis,cmc,moco,simclr} and masked image modeling~\cite{beit,simmim,mae,maskfeat,msn,ibot,data2vec,eva}. Contrastive learning promotes invariance~\cite{infomin} for two views of the same image and pushes apart representations for different images~\cite{uniformity} (or only invariance~\cite{byol,dino}); the resulting representations yield strong performance for linear or zero-shot transfer. Masked image modeling reconstructs the pixels~\cite{mae,simmim} or local features~\cite{data2vec}, often producing excellent fine-tuning transfer performance, especially in dense prediction tasks~\cite{mae}. The state of the art DINO v2~\cite{dinov2} leverages both approaches, and our approach shares a similar spirit.

\noindent\textbf{Supervised learning}~\cite{alexnet,vgg,resnet} used to be the dominant approach for learning transferable visual representations for various tasks~\cite{rcnn,decaf,sharif2014cnn}. Recent studies~\cite{li2021benchmarking,he2019rethinking} has shown that, the transferability of representations learned in this way is limited, \eg,  pre-training has no improvement over random initialization for dense prediction tasks (\eg, object detection) when the fine-tuning is long enough. Such limitation continues when the model has been scaled up to 22B~\cite{vit22b}. An alternative paradigm learns visual representations from text supervision~\cite{clip,align}, \eg, CLIP~\cite{clip}. This approach is more flexible (\ie, not requiring classes) and provides richer supervision, often learning generalizable representations.

\noindent\textbf{Generative models as representation learners.} A number of papers have explored the representations that are learned by generative models for various recognition tasks~\cite{donahue2019large,li2023mage}. As might be expected intuitively, such models indeed learn especially good representations for dense tasks, such as optical flow estimation~\cite{saxena2023surprising}, semantic segmentation~\cite{brempong2022denoising,xu2023open}, and depth estimation~\cite{zhao2023unleashing}. Another line of work~\cite{li2023your,clark2023text} adapt pre-trained diffusion models for zero-shot image recognition via analysis-by-synthesis. These approaches may need to be adapted when the architectures of the generative models change or a new family of generative model emerge. Our approach treats images as universal interfaces with the hope of better generality.

\noindent\textbf{Learning from synthetic data from generative models.} Synthetic data has been explored to train machine learning models in various domains~\cite{silver2017mastering,rosenberg2019speech,rossenbach2020generating,mimura2018leveraging,kumar2020data,taori2023alpaca,yang2020generative,meng2022generating,laclip}. 
In computer vision, the utilization of synthetic data for training models is common, ranging from optical flow~\cite{mayer2016large} and autonomous driving~\cite{abu2018augmented} to semantic segmentation~\cite{chen2019learning} and human pose estimation~\cite{varol2017learning}. Others~\cite{liu2022palm,jahanian2021generative} have explored synthetic data for representation learning, with the predominant approach of altering the latent variables of deep generative models. Our approach aligns with this research paradigm, but it diverges in its use of text-to-image models, which have also been investigated by other researchers~\cite{fake,zhou2023training,he2022synthetic}. But they use synthetic data for supervised learning~\cite{fake,scaling}. The closet work is StableRep~\cite{stablerep}, which also conducts representation learning but still needs a real text dataset.

\section{Approach}

In this paper, we study the problem of learning a visual encoder $f$ in the absence of real images or textual data. Our approach hinges on the utilization of three key resources: a language generation model ($g_1$), a text-to-image generative model ($g_2$), and a curated list of visual concepts ($C$). Our exploration include three steps: (1) we employ $g_1$ to synthesize a comprehensive set of image descriptions $T$, which encompass the range of visual concepts in $C$; (2) for each caption in $T$, we generate multiple images using $g_2$, culminating in an extensive synthetic image dataset $X$; (3) we train on $X$ to obtain a visual representation encoder $f$.

We use Llama-2 7B~\cite{llama2} and Stable Diffusion 1.5~\cite{stablediffusion} as $g_1$ and $g_2$, respectively, because of their fast inference speed. We anticipate that better $g_1$ and $g_2$ in the future will further enhance the effectiveness of this approach.

\subsection{Synthesizing captions}\label{sec:synthesize_caption}
To harness the capability of powerful text-to-image models for generating a substantial dataset of training images, we initially require a collection of captions that not only precisely depict an image but also exhibit diversity to encompass a broad spectrum of visual concepts.

We have developed a scalable approach to create such a large collection of captions, leveraging the in-context learning capability of LLMs~\cite{brown2020language}. Our method involves crafting specific prompt engineering templates that guide the LLM to produce the required captions. We start by gathering the concept list $C$ from some existing datasets, such as ImageNet-21k~\cite{imagenet} and Places-365~\cite{places}. For each concept $c \in C$, we consider three straightforward templates to generate captions effectively.

\begin{table*}[h]
\centering
\small
\begin{tabular}{ p{2.5cm}p{13.5cm} }
\toprule
Templates & In context examples \\
\hline
\emph{\textcolor{darkred}{$c$} --> caption}        & \textcolor{darkred}{revolver} --> Multiple antique \textcolor{darkred}{revolvers} lie on a wooden table, gleaming under soft, ambient light. \\
& \textcolor{darkred}{closet} --> The compact \textcolor{darkred}{closet}, 
brimming with clothes and shoes, exudes a feeling of organization. \\
& \textcolor{darkred}{zebra} --> A \textcolor{darkred}{zebra} is gallantly trotting across the vast, sunlit plains of the African savannah, creating a captivating black and white spectacle. \\
& \textcolor{darkred}{bus station} --> The bustling \textcolor{darkred}{bus station} thrums with restless energy, as travelers navigate through the crowded space, awaiting their journeys amid the echoes of departing buses. \\
\emph{\textcolor{darkred}{$c$}$, $\textcolor{darkblue}{$bg$} --> caption}    & \textcolor{darkred}{tiger}, \textcolor{darkblue}{forest} --> Two \textcolor{darkred}{tigers} are running together in the \textcolor{darkblue}{forest}. \\
& \textcolor{darkred}{lighter}, \textcolor{darkblue}{motorhome} --> In the cozy, cluttered environment of a well-traveled \textcolor{darkblue}{motorhome}, a sleek silver \textcolor{darkred}{lighter} holds dominion on the rustic wooden table. \\
& \textcolor{darkred}{sunset}, \textcolor{darkblue}{lake} --> Golden \textcolor{darkred}{sunset} hues reflect on a calm \textcolor{darkblue}{lake}, silhouetting a lone canoeist against a backdrop of fiery clouds. \\
\emph{\textcolor{darkred}{$c$}$, $\textcolor{darkgreen}{$rel$} --> caption}   & \textcolor{darkred}{kit fox}, \textcolor{darkgreen}{in front of} --> A group of small, fluffy, golden \textcolor{darkred}{kit foxes} is playfully gathered \textcolor{darkgreen}{in front of} a lush, green, towering forest backdrop. \\
& \textcolor{darkred}{cabbage}, \textcolor{darkgreen}{besides} --> A vibrant image portrays a lush, green \textcolor{darkred}{cabbage}, glistening with dewdrops, nestled \textcolor{darkgreen}{besides} a rustic, wooden crate full of freshly harvested vegetables. \\
\bottomrule
\end{tabular}
\caption{We show examples for the three synthesis templates. Such examples are used as demonstrations for Llama-2 to perform the in-context learning task. We have 176 such examples in total. Most of them are generated by prompting GPT-4~\cite{gpt4}, while a handful of others are human generated (in a 10M scale pilot study of synthetic captions, we do not notice significant differences between including or excluding human generated examples.)}
\label{tab:in_context_example}
\vspace{-5pt}
\end{table*}

\begin{itemize}
\item \emph{$c$ --> caption}. As the most direct and simple approach, we have the Llama-2 model sample a sentence for the concept $c$.

\item \emph{$c$, $bg$ --> caption}. We combine the visual concept $c$ with a background or setting $bg$. A naïve approach would randomly select both $c$ and $bg$, where $bg$ may correspond to a class name from a places dataset like \cite{places}. However, this method often leads to unlikely combinations in the real world, such as a blue whale in a football field. Our ablation experiments demonstrate that this strategy results in suboptimal performance, likely because the generated captions fall far outside the training distribution of $g_2$.  Instead, we employ GPT-4~\cite{gpt4} to generate a list of suitable backgrounds for the chosen concepts. This approach increases the likelihood of generating more plausible combinations, such as a tiger in a forest or a cat in a kitchen, enhancing the overall quality of the results.

\item \emph{$c$, $rel$ --> caption}. Given a visual concept $c$, we consider pairing it with a positional relationship word, $rel$. Take for instance, if $c$ signifies \emph{cat} and 
$rel$ translates to \emph{in front of}, our objective is to prompt the LLM to create captions such as \emph{a cute yellow cat is enjoying the fish in front of the sofa}. To add variety, we have a selection of 10 different positional relationship words that we randomly choose from.
\end{itemize}

For each of the three templates, we have prepared multiple demonstration examples that serve as instructions for the LLM to complete the caption synthesis task. Table~\ref{tab:in_context_example} shows a couple of examples for each template. In total, we have 106 examples for \emph{$c$-->prompt}, 50 examples for \emph{$c, bg$-->prompt}, and 20 examples for \emph{$c, rel$-->prompt}. Such examples are mostly collected by prompting GPT-4, with a handful from human. In a pilot study, we do not observe difference between including or excluding human generated examples.

\begin{figure}[t]
\centering
\small
\includegraphics[width=1.0\linewidth]{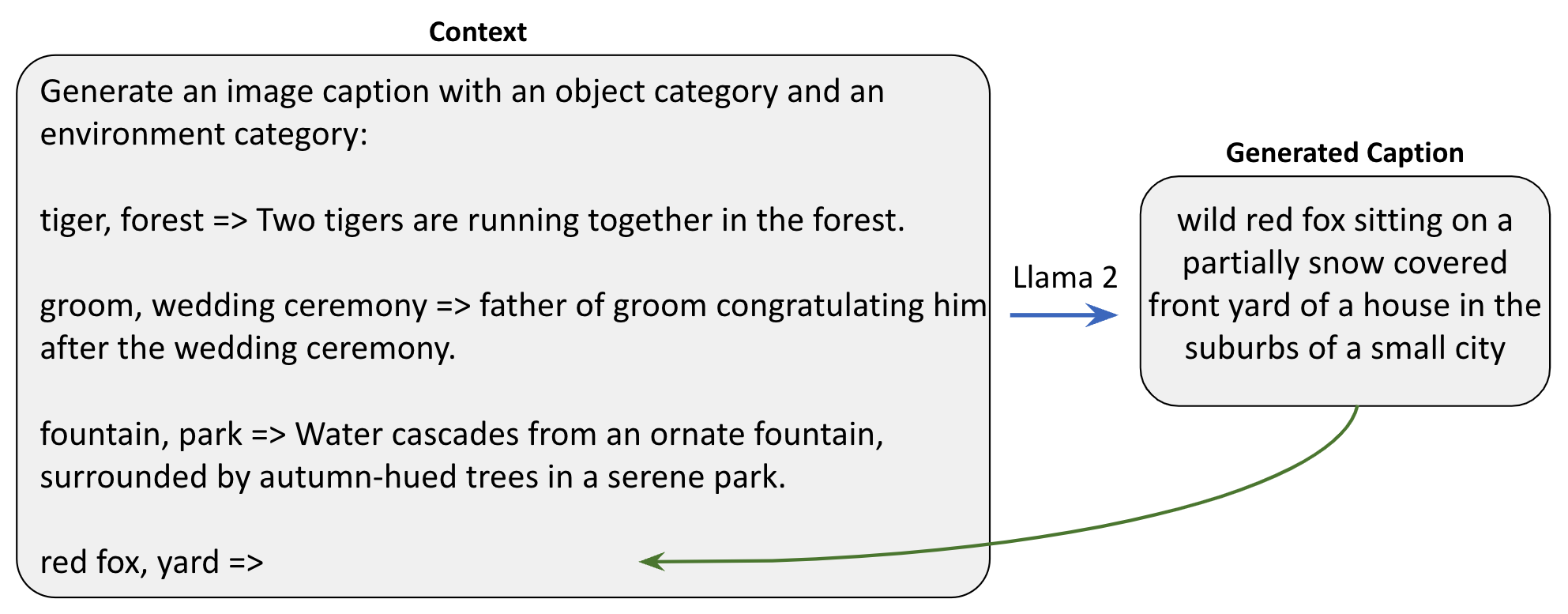}
\caption{In-context caption generation using Llama-2~\cite{llama2}. We randomly sample three in-context examples for each inference run.}
\label{fig:llama}
\vspace{-5pt}
\end{figure}

In the stage of generating captions in-context, we select a concept and one of the three templates. Next, we randomly pick three examples from the chosen template and frame the caption generation as a text completion task. This process is illustrated in Figure~\ref{fig:llama}.

\subsection{Synthesizing Images}

\begin{figure*}[t]
  \centering
    \includegraphics[width=\linewidth]{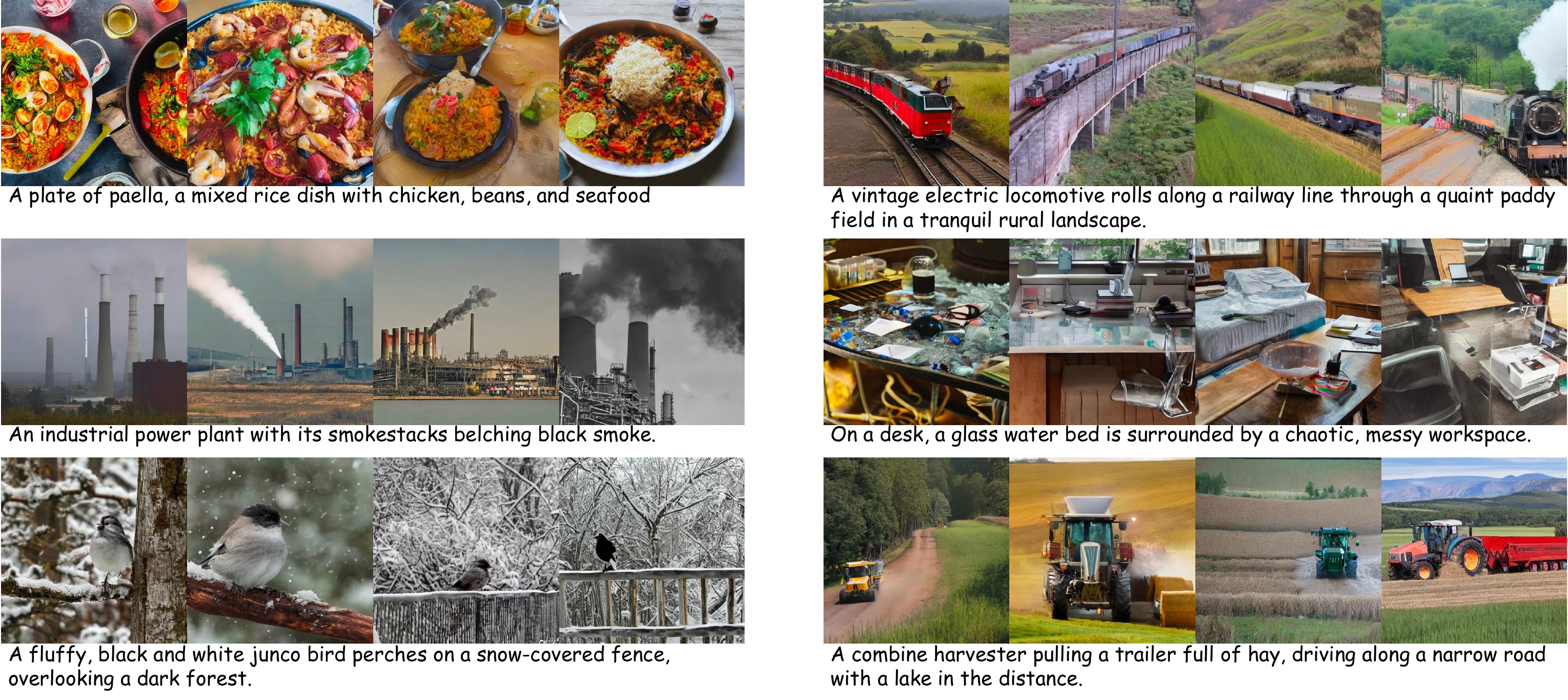}\\
   \vspace{-2.5mm}
  \caption{\small Random examples of synthetic captions and images generated in our~\name~pipeline. Each caption comes with 4 images.}
  \label{fig:visualization}
  \vspace{-5pt}
\end{figure*}

For each text caption, we generate a variety of images by initiating the reverse diffusion process with different random noise. The Classifier-Free Guidance (CFG) scale is a crucial factor in this process. A higher CFG scale enhances the quality of the samples and the alignment between text and image, whereas a lower scale results in more diverse samples and better adherence to the original conditional distribution of images based on the given text. Following the approach used in StableRep~\cite{stablerep}, we opt for a lower CFG scale, specifically 2.5, and produce 4 images for each caption. Examples of these images can be seen in Figure~\ref{fig:visualization}.

\subsection{Representation Learning}

Our representation learning method is built upon StableRep~\cite{stablerep}. The key component of our approach is the multi-positive contrastive learning loss~\cite{supcon} which works by aligning (in the embedding space) images generated from the same caption. We additionally combine multiple techniques from other self-supervised learning methods, including a patch-level masked image modeling objective. We briefly review StableRep and elaborate on the added modules.

\noindent\textbf{StableRep}~\cite{stablerep} minimizes the cross-entropy loss between a ground-truth assignment distribution and a contrastive assignment distribution. Consider an encoded anchor sample $\boldsymbol{a}$ and a set of encoded candidates $\{\boldsymbol{b}_1, \boldsymbol{b}_2, ..., \boldsymbol{b}_K\}$. The contrastive assignment distribution $\rvq$ describes how likely the model predicts $\boldsymbol{a}$ and each $\boldsymbol{b}$ to be generated from the same caption, and the ground-truth distribution is the actual match between $\boldsymbol{a}$ and $\boldsymbol{b}$ ($\boldsymbol{a}$ is allowed to match multiple $\boldsymbol{b}$):
\begin{align}
    \rvq_{i} = \frac{\exp( \boldsymbol{a} \cdot \boldsymbol{b}_i / \tau)}{\sum_{j=1}^{K}\exp(\boldsymbol{a} \cdot \boldsymbol{b}_j/\tau)}
    \label{eq:contrastive_distribution}
    \\
    \rvp_{i} = \frac{\mathbbm{1}_{\text{match}(\boldsymbol{a}, \boldsymbol{b}_i)}}{\sum_{j=1}^{K} \mathbbm{1}_{\text{match}(\boldsymbol{a}, \boldsymbol{b}_j)}}
\end{align}
where $\tau \in \mathcal{R}_{+}$ is the scalar temperature, $\boldsymbol{a}$ and all $\boldsymbol{b}$ have been $\ell_2$ normalized, and the indicator function $\mathbbm{1}_{\text{match}(\cdot,\cdot)}$ indicates whether two samples are from the same caption. The contrastive loss for $\boldsymbol{a}$ is given as
\begin{align}\label{eq:contrastive}
    \mathcal{L}(\boldsymbol{a}) = H(\rvp, \rvq) = - \sum_{i=1}^{K} \rvp_{i} \log\rvq_{i}
\end{align}

\noindent\textbf{iBOT}~\cite{ibot} is a masked image modeling objective, wherein a localized patch is masked, and the model is tasked with predicting the tokenized representation of said masked patch.
It adapts the DINO~\cite{dino} objective from the image level into the patch level. We follow~\cite{ruan2022weighted} to replace the softmax-centering method with the iterative Sinkhorn-Knopp (SK) algorithm~\cite{sinkhorn}. We run SK for 3 iterations to build the prediction target.

\noindent\textbf{Exponential Moving Average (EMA)} is firstly introduced into self-supervised learning by MoCo~\cite{moco}. We use EMA to encode crops as $\boldsymbol{b}$ and to produce the targets for iBOT loss. We update the EMA model as $\theta_{ema} \leftarrow \lambda \theta_{ema} + (1-\lambda) \theta$, following a cosine schedule for $\lambda$ from 0.994 to 1 during training~\cite{byol,dinov2}. We find the EMA module not only increases the final performance, but also improves the training stability for long training schedules.

\noindent\textbf{Multi-crop} strategy is introduced by~\cite{swav} as a smart way to improve computation efficiency, and is adopted in this paper. For these local crops, we only employ the contrastive loss, omitting the iBOT loss. Local crops are encoded only by the student network, and matched to global crops from the same caption encoded by the EMA model. Such reuse of global crops saves computation. 
For each image $x$, where we generate a single global crop $x^g$ alongside $n$ local crops $x^l$, the final loss can be expressed as follows:
\begin{align}\label{eq:total_loss}
    \mathcal{L}(x^g) + \frac{1}{n}\sum_{i=1}^{n}\mathcal{L}(x^l_i) + \mathcal{L}^{iBOT}(x^g)
\end{align}

\subsection{Implementation}

\noindent\textbf{Concept list.} We concatenate class names from various datasets, including IN-1k~\cite{imagenet}, IN-21k (we keep the most frequent 13k classes), Aircraft~\cite{aircraft}, Cars~\cite{cars}, DTD~\cite{dtd}, Flowers~\cite{flowers}, Pets~\cite{pets}, Sun397~\cite{sun397}, Caltech-101~\cite{caltech101}, Food-101~\cite{food101}, and Places-365~\cite{places}. If the concept is a place (\ie SUN397 and Places) or a texture (\ie DTD), we only apply the \emph{$c$ --> caption} template. For fine-grained classes such as pets or flowers, we employ GPT-4 to generate a consolidated list of probable backgrounds, rather than producing distinct lists for each specific class. We  favor more frequent sampling from IN-1k, Food101, Cars, Aircraft, and Flowers.

\noindent\textbf{Batches.} For each training batch, we sample 2048 captions (except when noted), and use all of the 4 images generated by each caption. We generate 1 global and 4 local crops for each image. As a result, each batch contains 8192 global crops, which is similar with prior work~\cite{simclr,byol,mocov3,stablerep}.

\noindent\textbf{Masking.} For the iBOT loss, we randomly choose $50\%$ images inside a batch to mask, and randomly mask $50\%$ of the tokens in each chosen image. We use 65536 prototypes. While the target from the EMA model is ascertained using the SK algorithm, we apply softmax normalization to the output of the student model.

\noindent\textbf{Projection heads.} We follow the design in MoCo v3~\cite{mocov3} and DINO~\cite{dino} for the contrastive and iBOT loss heads, respectively, ensuring consistency with established methods.  

\noindent\textbf{Other hyper-parameters.} 
We set the temperature in the contrastive loss to $0.08$. For the temperature used in the iBOT loss, we linearly increase it from 0.04 to 0.07 over 4000 iterations, and keep it as 0.07 afterwards, as in DINO~\cite{dino}. Additionally, the weight decay parameter is incrementally adjusted from 0.04 to 0.2, adhering to a cosine schedule.
\section{Experiment}

We first perform an ablation study to evaluate the efficacy of various designs and modules within our pipeline. Then we proceed to scale up the volume of synthetic data.

\subsection{Study different components}

We analyze each component of~\name, and ablate their effectiveness in two measurements: (1) linear probing performance on IN-1k; (2) average accuracy of linear transfer on fine-grained datasets Aircraft~\cite{aircraft}, Cars~\cite{cars}, DTD~\cite{dtd}, Flowers~\cite{flowers}, Pets~\cite{pets}, Sun397~\cite{sun397}, Caltech-101~\cite{caltech101}, Food-101~\cite{food101}, and Pascal VOC~\cite{pascalvoc}. For analysis conducted in this subsection, we train ViT-B/16~\cite{vit} models for 85000 iterations, and use the \texttt{cls} token as image representation.

\begin{table}[t]
\vspace{-10pt}
\setlength{\tabcolsep}{6pt}
\begin{center}
\begin{small}
\begin{tabular}{lcccc}
\toprule
\multirow{2}{*}{captions} & \multicolumn{2}{c}{StableRep} & \multicolumn{2}{c}{SynCLR}\\

 & IN & avg. & IN & avg. \\
\midrule
\textcolor{gray!80}{cc12m} & \textcolor{gray!80}{73.0} & \textcolor{gray!80}{81.6} & \textcolor{gray!80}{77.1} & \textcolor{gray!80}{85.3} \\ 
IN+h+Places        & 75.4 & 80.0 & 78.7 & 83.0 \\ 
IN+Places+LLM    & 73.7 & 76.9 & 77.6 & 81.8 \\ 
IN+OurBG+LLM    & 75.3 & 78.5 & 78.2 & 81.9 \\ 
\midrule
our final config. & \textbf{75.8} & \textbf{85.7} & \textbf{78.8} & \textbf{88.1} \\ 
\bottomrule

\end{tabular}
\vspace{-5pt}
\caption{\small \textbf{Comparison of different caption synthesis strategies}. We report top-1 ImageNet linear evaluation accuracy and the average accuracy over 9 fine-grained datasets. Every item here includes 10M captions and 4 images per caption.}
\label{tab:caption_results}
\vspace{-10pt}
\end{small}
\end{center}
\end{table}
\begin{table}[t]
\begin{center}
\begin{small}
  \begin{tabular}{l c c c}
    \toprule
	  CFG & 2 & 3 & 4 \\
    \midrule
	  IN top-1 & 72.8 & 72.6 & 72.6  \\
    \bottomrule
\phantom{latex!}
  \end{tabular}
\vspace{-15pt}
\caption{\small \textbf{Classifier-free guidance scale (CFG).}
Contrastive loss prefers small CFG scale but is not very sensitive to it.
}
\label{tab:cfg}
\end{small}
\end{center}
\vspace{-20pt}
\end{table}

\noindent\textbf{Synthesize captions.} Following~\cite{stablerep}, we use cc12m~\cite{cc12m} real captions as our baseline, which has 10M sentences. To synthesize captions, we design the following variants: (a) \emph{IN+h+Places} randomly combines one IN class plus its hypernyms in WordNet graph, with one place class; (b) \emph{IN+Places+LLM} uses the \emph{$c, bg$ --> caption} in-context synthesis template with $c$ from IN and $bg$ from places; (c) \emph{IN+ourBG+LLM} uses the background classes output by GPT-4, instead of Places; (d) \emph{ours} means our full configuration specified in Section~\ref{sec:synthesize_caption}. 
For each of the config, we generate 10M captions. If not enough, we do duplication.

Results are summarized in Table~\ref{tab:caption_results}, where we train both StableRep and SynCLR to avoid biases favored by a single method. Compared to a real caption dataset cc12m, simply concatenating IN and Places class names improves the ImageNet linear accuracy but reduces the fine-grained classification performance. Interestingly, naively asking Llama to combine IN and Places classes into captions yields the worst performance. Replacing random background from places with GPT generated background improves the accuracy. This shows the importance of synthesizing captions that follow the distribution of real captions, which were used to train the text-to-image model. Finally, our full configuration achieves the best accuracy on both ImageNet and fine-grained classification. Another advantage of our synthesis method is its scalability -- scale up to hundreds of millions of captions with little duplication. In contrast, if we concatenate IN classes with Places classes, there are at most 365k unique captions.

\noindent\textbf{Synthesize images.} There are two major parameters in this process: number of images per caption and classifier free guidance scale. For the former, we find generating 4 images is almost able to reproduce StableRep~\cite{stablerep}'s performance (10 images) when using cc12m captions (ours 73.0$\%$ v.s. StableRep 73.5$\%$ on ImageNet). Thus we stick to 4. For guidance scale, we briefly find the contrastive loss is not very sensitive to CFG in a pilot study, as shown in Table~\ref{tab:cfg}. Thus we stick to 2.5, similar as StableRep~\cite{stablerep}.
\begin{table}[t]
\vspace{-10pt}
\setlength{\tabcolsep}{5.5pt}
\begin{center}
\begin{small}
\begin{tabular}{lcccccc}
\toprule

method & EMA & iBOT & MC & IN & avg. & ADE20k \\
\midrule
StableRep   &  &  &  & 75.8 & 85.7 & - \\
\midrule
     & \checkmark & & & 76.7 & 86.7 & 48.0 \\
     & \checkmark & \checkmark & & 77.6 & 87.1 & 50.5 \\
     & \checkmark & & \checkmark & 78.6 & 87.8 & 49.5\\
\name & \checkmark & \checkmark & \checkmark & 78.8 & 88.1 & 50.8\\
\bottomrule

\end{tabular}
\vspace{-5pt}
\caption{\textbf{Important components for our model.} ViT-B/16 models are trained for 85000 iterations. We study the modules that affect the ImageNet linear evaluation, the fine-grained classification (avg.), and ADE20k segmentation. 
}
\label{tab:model_results}
\end{small}
\end{center}
\vspace{-10pt}
\end{table}
\begin{table}[t]
\setlength{\tabcolsep}{10pt}
\begin{center}
\begin{small}
\begin{tabular}{lcc}
\toprule

method & IN & avg. \\
\midrule
Supervised CE          & 71.9 & 75.0\\
SimCLR                 & 63.6 & 67.9 \\
\midrule
\name                   & \textbf{75.3} & \textbf{78.5} \\
\bottomrule

\end{tabular}
\caption{\textbf{Comparison of different learning objectives.} These objectives assume different level of classification granularity, as shown in Figure~\ref{fig:model_diff}. Our modeling, \ie, defining classes as captions, outperforms the other two. To accomondate Supervised CE training, all items here used \emph{IN+OurBG+LLM} entry in Table~\ref{tab:caption_results}.}
\label{tab:diff_objective}
\vspace{-15pt}
\end{small}
\end{center}
\end{table}
\begin{table*}[th]
  \vspace{-10pt}
  \setlength{\extrarowheight}{5pt}
  \renewcommand{\arraystretch}{0.75}
  \centering
  \small
  \begin{tabularx}{\linewidth}{p{1.4cm}cccc|c|cccccccccc}
    & text & img & \# imgs & & \datatagnew{\textbf{ImageNet}} & \datatagnew{Aircraft} & \datatagnew{Cars} &  \datatagnew{DTD} &  \datatagnew{Flowers}  & \datatagnew{Pets} & \datatagnew{SUN397}  &  \datatagnew{Caltech-101} &  \datatagnew{Food-101} &  \datatagnew{VOC2007} & \datatagnew{\textbf{Average}} \\
    
    \shline
    StableRep & real & syn & 100M & ViT-B/16 & 75.7 & 59.2 & 83.5 & 80.1 & 97.3 & 88.3 & 74.3 & 94.7 & 85.1 & 87.9 & 83.4 \\ [4pt]

    \multirow{2}{=}{CLIP} 
    & \multirow{2}{*}{real} & \multirow{2}{*}{real} & \multirow{2}{*}{400M} & ViT-B/16 & 80.2 & 59.5 & 86.7 & 79.2 & 98.1 & 93.1 & 78.4 & 94.7 & 92.8 & 89.2 & 85.7 \\
    & & & & ViT-L/14 & 83.9 & 69.4 & 90.9 & 82.1 & 99.2 & 95.1 & 81.8 & 96.5 & 95.2 & 89.6 & 88.9 \\ [4pt]

    \multirow{3}{=}{OpenCLIP} 
    & \multirow{3}{*}{real} & \multirow{3}{*}{real} & 400M & ViT-B/16 & 78.9 & 61.1 & 92.3 & 81.9 & 98.2 & 91.5 & 77.9 & 95.2 & 90.9 & 88.0 & 86.3 \\
    & & & 400M & ViT-L/14 & 82.3 & 67.1 & 94.0 & 83.6 & 98.8 & 92.5 & 81.0 & 96.4 & 93.4 & 88.8 & 88.4 \\ 
    & & & 2B & ViT-L/14 & 83.4 & 71.7 & 95.3 & 85.3 & 99.0 & 94.2 & 82.2 & 97.5 & 94.1 & 88.9 & 89.8 \\ [4pt]

    \multirow{2}{=}{DINO v2*} 
    & \multirow{2}{*}{-} & \multirow{2}{*}{real} & \multirow{2}{*}{142M} & ViT-B/14 & \textbf{83.9}\(^{\dag}\) & 79.4 & 88.2 & 83.3 & 99.6 & 96.2 & 77.3 & 96.1 & 92.8 & 88.2 & \textbf{89.0} \\
    & & & & ViT-L/14 & \textbf{85.7}\(^{\dag}\) & 81.5 & 90.1 & 84.0 & 99.7 & 96.6 & 78.7 & 97.5 & 94.3 & 88.3 & 90.1 \\ [4pt]

    \hline
    \multirow{2}{=}{SynCLR} 
    & \multirow{2}{*}{syn} & \multirow{2}{*}{syn} & \multirow{2}{*}{600M} & ViT-B/16 & 80.7 & 81.7 & 93.8 & 79.9 & 99.1 & 93.6 & 76.2 & 95.3 & 91.6 & 89.4 & \textbf{89.0} \\
    & & & & ViT-L/14 & 83.0 & 85.6 & 94.2 & 82.1 & 99.2 & 94.1 & 78.4 & 96.1 & 93.4 & 90.3 & \textbf{90.4} \\
    
  \end{tabularx}
  \caption{
  \small
  \textbf{Comparison on ImageNet linear evaluation and fine-grained classificaton.} \name~achieves comparable results with OpenAI's CLIP and DINO v2 models, despite \emph{only} using synthetic data. *DINO v2 modes are distilled from a ViT-g model, thus advantageous in this comparison. \(^{\dag}\) we rerun only using \texttt{cls} token instead of concatenating multiple layers presented in the original DINO v2 paper~\cite{dinov2}. 
  }
  \label{tab:compare}
\vspace{-10pt}
\end{table*}

\noindent\textbf{Model components.} We present the improvement of accuracy brought by different modules in Table~\ref{tab:model_results}. Compared to the baseline StableRep, adding a teacher EMA model improves the IN linear accuracy by 0.9$\%$. Further adding iBOT local objective or the multi-crop strategy increases the accuracy by 0.9$\%$ and 1.9$\%$, respectively. Combining all of them results in our full \name~model, which achieves 78.8$\%$ top-1 IN linear accuracy. The fine-grained classification performance follows a similar trend, and reaches 88.1$\%$. Besides, we test the transfer ability to semantic segmentation on ADE20k. The iBOT objective brings 1.0 more mIoU than multi-crop strategy, demonstrating the effectiveness of masked image modeling for dense prediction tasks.

\noindent\textbf{Compare to SimCLR and supervised training. } We compare the three different representation learning objectives shown in Figure~\ref{fig:model_diff}, which classify images at different levels of granularity. Since supervised cross-entropy training requires a fixed set of balanced classes (indeed both \emph{fixed set} and \emph{balance} are limitations of such method), we use the \emph{IN+ourBG+LLM} configuration where we have 1000 balanced classes (\ie, each class has 40k images). The supervised training recipe follows~\cite{augreg}. For a fair comparison with SimCLR, we remove all unmatched modules (\ie, EMA, iBOT, and MC) to make sure that the only difference between SimCLR and our \name~is the classification granularity defined by the contrastive loss. For all of them, we do pre-training and then linear probing on the target dataset.

Table~\ref{tab:diff_objective} presents the comparison. Our multi-positive objective, which defines images as the same class if they are generated by the same caption, achieves the best performance. It outperforms supervised cross-entropy training and SimCLR by 3.4$\%$ and 11.7$\%$ for top-1 accuracy on ImageNet linear evaluation, and by 3.5$\%$ and 10.6$\%$ on fine-grained classification tasks. Besides, our objective does not require balance between samples from a fixed set of classes, making it easier to scale up.

\subsection{Scaling up}

After we have ablated different components, we scale up our experiments. Specifically, we synthesize a dataset of 150M captions, called \emph{SynCaps-150M}, from which we generate 600M images. We train both ViT-B/16 and ViT-L/14 (no SwiGLU~\cite{shazeer2020glu} or LayerScale~\cite{touvron2021going}), and extend the training schedules to 500k steps with a batch size of 8192 captions. We use 224x224 resolution for all pre-training tasks.

We compare \name~with OpenAI's CLIP~\cite{clip}, OpenCLIP~\cite{openclip}, and DINO v2~\cite{dinov2}, which represent \emph{learning from data}. We note that ViT-B/14 and ViT-L/14 from DINO v2 are distilled from a ViT-g~\cite{vitscale} model, which makes DINO v2 advantageous in our comparison. We also includes StableRep~\cite{stablerep}, which uses the \emph{hybrid} paradigm.

\noindent\textbf{ImageNet linear evaluation.} For fair comparison, \texttt{cls} token from the last block is used as representation across all models (whereas in DINO v2, results are from concatenating multiple layers). As shown in Table~\ref{tab:compare}, \name~achieves 80.7$\%$ with ViT-B and 83.0$\%$ with ViT-L. This is similar as CLIP, but still lags behind DINO v2 by 3.2$\%$ and 2.7$\%$, respectively, partially because of the extra distillation in DINO v2. We note~\name~has already outperformed other self-supervised methods pre-trained directly on ImageNet-1k (\eg, DINO achieves 78.2$\%$ with ViT-B/16 and iBOT reaches 81.0$\%$ with ViT-L/16).

\noindent\textbf{Fine-grained classification.} On the nine fine-grained datasets we have evaluated in Table~\ref{tab:compare}, \name~achieves very similar average accuracy as DINO v2, \eg, 89.0$\%$ v.s. 89.0$\%$ for ViT-B, and 90.1$\%$ vs 90.4$\%$ for ViT-L. Both \name~and DINO v2 have curated the pre-training data to include the distribution for these datasets (but in different ways and portions), and end up with similar performance. Interestingly, \name~outperforms others on Aircraft and Cars, possibly because we favor more frequent sampling towards them.
This can be an advantage for synthetic data when we know what downstream tasks to solve. Besides, \name~outperforms CLIP and StableRep by 3.3$\%$ and by 5.6$\%$ for ViT-B, respectively.

\begin{table}[t]
\centering
\small
\setlength{\tabcolsep}{4.5pt}
\begin{tabular}{llcll}
\toprule
method & pre-train data & distill & ViT-B & ViT-L \\
\midrule
StableRep    & hybrid, 100M   & & 49.4   & - \\
MoCo v3      & real, IN1K-1M & & 47.3 & 49.1 \\
BEiT         & real, IN1K-1M+DALLE & & 47.1 & 53.3 \\
MAE          & real, IN1K-1M & & 48.1 & 53.6 \\
iBOT         & real, IN1K-1M & & 50.0 & - \\
CLIP         & real, WIT-400M & & 52.6 & - \\
BEiT v2      & real, WIT-400M, IN1K & \checkmark & 53.1 & 56.7 \\
DINO v2      & real, LVD-142M & \checkmark & \textbf{54.4} \(^{\dag}\) & 57.5\(^{\dag}\) \\
\midrule
\name        & synthetic, 600M & & \textbf{54.3} & \textbf{57.7} \(^{\dag}\) \\
\bottomrule
\end{tabular}
\caption{\textbf{ADE20K semantic segmentation} (mIoU) using UperNet, with single scale at 512x512 resolution. \(^{\dag}\) use patch size of 14x14, thus adapt to 518x518 resolution.
}
\label{tab:ade20k} 
\end{table}

\noindent\textbf{Semantic segmentation.} To evaluate the pixel-level understanding ability of \name, we fine-tune the pre-trained models on ADE20k~\cite{ade20k}, following the setup in~\cite{mae,beit}. UperNet~\cite{upernet} is used as the task layer, and we evaluate with a single-scale, \ie 512x512. Besides CLIP and DINO v2, we also compare to self-supervised methods pre-trained on ImageNet, as well as BEiT v2~\cite{beitv2}, which distills from CLIP. Table~\ref{tab:ade20k} shows that our \name~outperforms self-supervised methods trained on IN-1k by a clear marge, \eg, 4.3 higher mIoU than iBOT. Despite not involving a high resolution pre-training period like DINO v2 (\eg, 518x518), \name~performs similarly with DINO v2 (0.1 lower for ViT-B possibly because DINO v2 uses a smaller patch size of 14x14, but 0.2 higher for ViT-L). This suggests~\name~pre-training is suitable for dense prediction tasks.

\begin{table}[t]
\centering
\small
\setlength{\tabcolsep}{6pt}
\begin{tabular}{llcc}
\toprule
method & pre-train data & ViT-B & ViT-L \\
\midrule
MoCo v3      & real, IN1K-1M & 83.2 & 84.1 \\
SimMIM       & real, IN1K-1M & 83.8 &  - \\
MAE          & real, IN1K-1M & 83.6 & 85.9 \\
PeCo         & real, IN1K-1M & 83.6 & 85.9 \\
data2vec     & real, IN1K-1M & 84.2 & 86.6 \\
iBOT         & real, IN21K-14M & 84.4 & 86.6 \\
BEiT v2      & real, WIT-400M+IN1k-1M & 85.5 & 87.3 \\
CLIP         & real, WIT-400M & 85.2 & 87.5\(^{\dag}\) \\
\multirow{2}{*}{OpenCLIP}  & real, LAION-400M & 85.0 & 86.6\(^{\dag}\) \\
             & real, LAION-2B & - & 87.1\(^{\dag}\) \\
\midrule
\name        & synthetic, 600M & \textbf{85.8} &  \textbf{87.9}\(^{\dag}\) \\
\bottomrule
\end{tabular}
\caption{\textbf{Top-1 accuracy on ImageNet with fine-tuning evaluation.}. Models are fine-tuned at 224x224 resolution. \(^{\dag}\) use patch size of 14x14.}
\label{tab:imagenet_finetune} 
\vspace{-10pt}
\end{table}

\noindent\textbf{ImageNet fine-tuning.} We evaluate the fine-tuning transfer ability of \name~on ImageNet. We compare with other state of the art self-supervised methods~\cite{mocov3,beit,simmim,mae,peco,data2vec,ibot} in Table~\ref{tab:imagenet_finetune}. Our \name~outperforms models trained on ImageNet images or large scale image datasets. 
Specifically, \name~outperforms OpenCLIP ViT-L trained on Laion-2B, which is the dataset Stable Diffusion (the text2image model we used) is trained on. 
This contrasts with~\cite{fake,scaling}, which shows that directly training a classifier on synthetic images yields bad classification accuracy. Our finding suggests synthetic images are good for training representations, which later can be easily adapted to a downstream task with limited amount of real data.

\subsection{Further analysis}

\name~requires a list of concepts $C$ to start off. But how will \name~transfer to concepts outside our list?

\begin{table}[t]
\setlength{\tabcolsep}{3.5pt}
\setlength{\extrarowheight}{5pt}
\renewcommand{\arraystretch}{0.75}
\begin{center}
\begin{small}
\begin{tabular}{p{1.2cm}c|cccccc|c}
& & \datatagnew{\textbf{EuroSAT}} & \datatagnew{GTSRB} & \datatagnew{Country211} & \datatagnew{MNIST} & \datatagnew{RESISC45} & \datatagnew{KITTI} & \datatagnew{\textbf{Average}} \\
\shline
\multirow{2}{=}{CLIP} & \footnotesize ViT-B/16 & 97.1 & 86.6 & 33.3 & 99.0 & 92.7 & 64.7 & \textbf{78.9} \\
& \footnotesize ViT-L/14 & 98.2 & 92.5 & 42.9 & 99.2 & 94.1 & 69.2 & \textbf{82.7} \\
\multirow{2}{=}{DINO v2} & \footnotesize ViT-B/14 & 96.0 & 72.8 & 21.6 & 98.6 & 92.5 & 75.3 & 76.1 \\
& \footnotesize ViT-L/14 & 96.7 & 74.1 & 24.1 & 98.2 & 93.8 & 76.9 & 77.3 \\
\multirow{2}{=}{\name} & \footnotesize ViT-B/16 & 96.6 & 78.6 & 21.0 & 98.4 & 93.7 & 77.3 & 77.6 \\
& \footnotesize ViT-L/14 & 96.7 & 79.2 & 24.3 & 98.5 & 93.8 & 78.0 & 78.4 \\
\bottomrule
\end{tabular}
\caption{\textbf{Generalization to concepts not seen by DINO v2 and \name.} \name~outperforms DINO v2. CLIP achieves the best accuracy, possibly because its training data includes similar concepts as these datasets.}
\label{tab:unseen}
\end{small}
\end{center}
\vspace{-10pt}
\end{table}

\begin{table}[t]
\setlength{\tabcolsep}{10pt}
\begin{center}
\begin{small}
\begin{tabular}{lcccc}
\toprule
 & \multicolumn{2}{c}{\name} & \multicolumn{2}{c}{CLIP} \\
 & IN & avg. & IN & avg. \\
\midrule
SynCaps-150M   & \textbf{80.7} & \textbf{89.0} & 78.3 & 87.7 \\
Laion-400M     & 78.9 & 86.5 & 76.6 & 84.9 \\
\bottomrule

\end{tabular}
\caption{\textbf{Compare \name~with CLIP on the same synthetic data.} We observe that: (1) \name~outperforms CLIP; (2) in our setup, \ie, generating 4 images per caption, SynCaps-150M yields better representations for both \name~and CLIP.}
\label{tab:syn_clip}
\end{small}
\end{center}
\vspace{-13pt}
\end{table}

\noindent\textbf{Generalize to unseen concepts.} We consider additional datasets whose classes are outside the synthesis list, including EuroSAT~\cite{eurosat}, GTSRB~\cite{gtsrb}, Country211~\cite{clip}, MNIST~\cite{mnist}, RESISC45~\cite{cheng2017remote}, and KITTI distances~\cite{kitti}. These datasets, except for KITTI, are also outside the curation list of DINO v2. Therefore, it is also a generalization test for DINO v2.
Table~\ref{tab:unseen} shows the linear probing results. \name~outperforms DINO v2 by 1.5$\%$ for ViT-B and 1.1$\%$ for ViT-L, respectively. This suggests the representations of \name~generalize. CLIP outperforms \name~and DINO v2, with most gains coming from Country211. An explanation is CLIP's training data contains similar country flags which are not in the training sets of \name~and DINO v2.

\begin{figure*}[ht]
\centering
\small
\includegraphics[width=\linewidth]{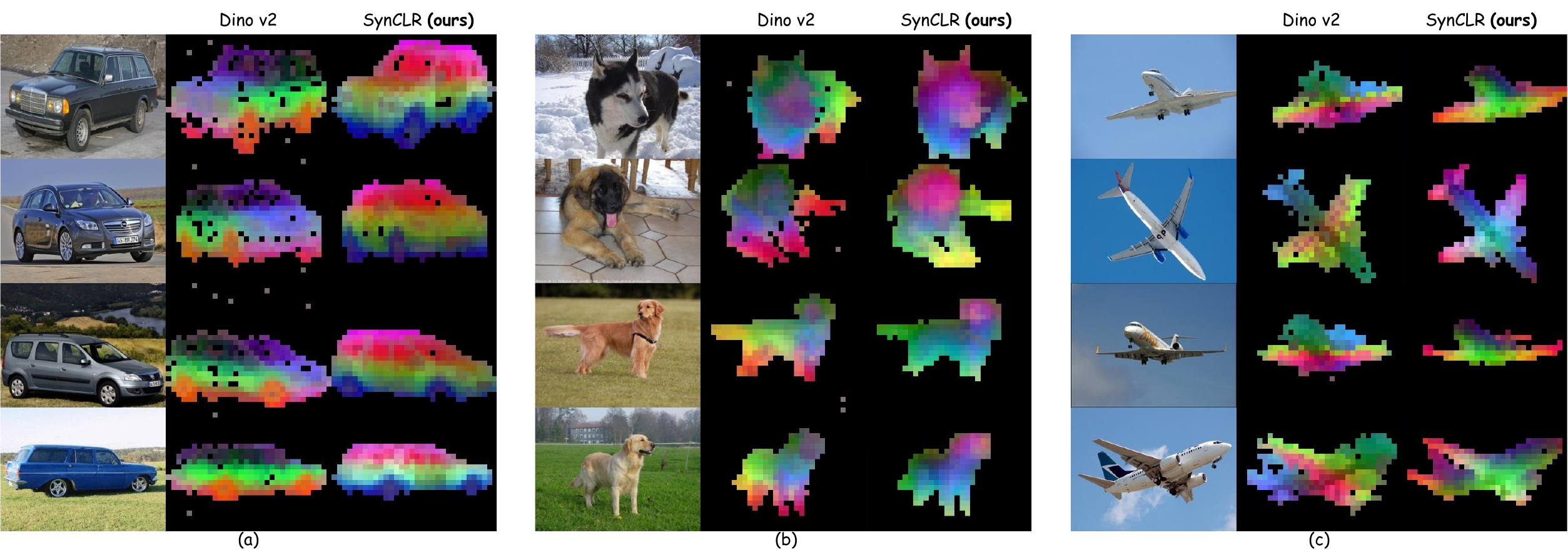}
\caption{\small \textbf{PCA visualization.} Follow DINO v2~\cite{dinov2}, we compute a PCA between the patches of the images from the same set and colorize by their first 3 components. Compared to DINO v2, \name~produces more accurate maps for cars (\eg, zoom-in to see the two bars on the roof of the first car, and the three side windows of the third car) and airplanes (\eg, the boundaries), while being slightly worse for dogs (\eg, heads). We use ViT-L/14 for both methods. Images are resized to 336x448 resolution before being fed into the networks, yielding 24x32 visualization grids.}
\label{fig:pca_visualization}
\end{figure*}

\begin{figure}[t]
\centering
\small
\includegraphics[width=0.85\linewidth]{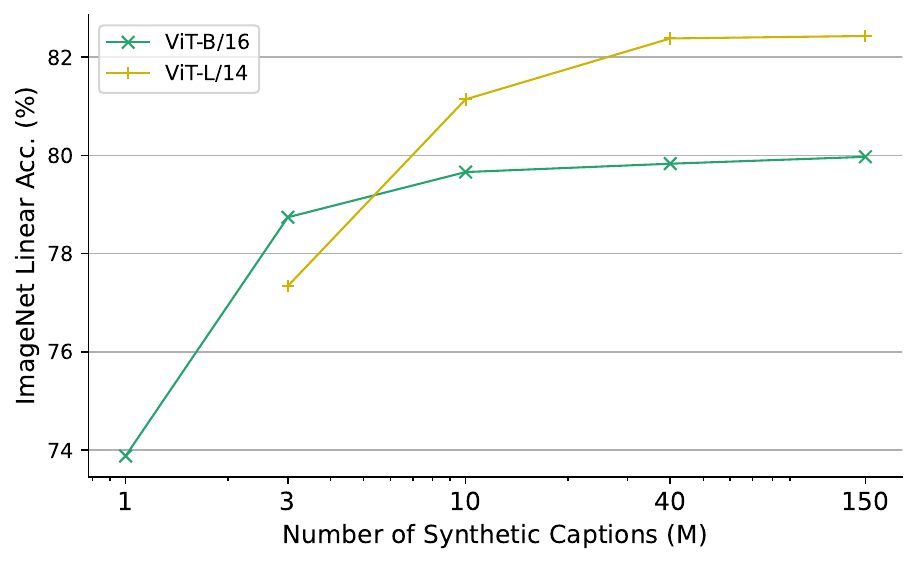}
\caption{\small\textbf{ImageNet linear} accuracy w/ different training scales.}
\label{fig:data_scale_imagenet}
\end{figure}

\begin{figure}[t]
\centering
\small
\includegraphics[width=0.85\linewidth]{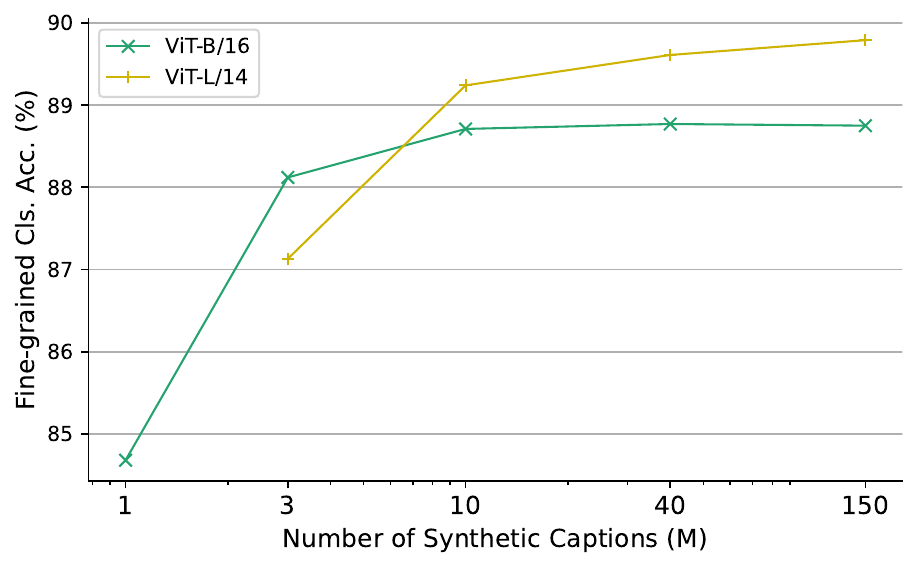}
\caption{\small\textbf{Fine-grained classification} w/ different training scales.}
\label{fig:data_scale_finegrained}
\vspace{-5pt}
\end{figure}

Given that both captions and images are synthesized, a natural question arises: how would CLIP training perform on such data?

\noindent\textbf{Compare to CLIP training.} We use the same data to train a ViT-B CLIP model. For each caption, we randomly choose 1 out of the 4 synthesized images in each iteration. Following common practice~\cite{clip}, we train for 32 epochs with a batch size of 32768. This model achieves 44.4$\%$ zero-shot accuracy on IN-1k. 
The \emph{SynCaps-150M} row in Table~\ref{tab:syn_clip} presents the linear probing results. Synthetic CLIP learns reasonably good features, reaching 78.3$\%$ on IN-1k and 87.7$\%$ on fine-grained datasets. However, \name~is still better.

We have also repeated our experiments with Laion-400M captions, \ie, generate 4 images for each caption and train \name~and CLIP. The comparison between rows \emph{SynCaps-150M} and \emph{Laion-400M} in Table~\ref{tab:syn_clip} suggests synthetic captions are also favorable on a large scale.

\noindent\textbf{PCA visualization.} %
Following the method used in DINO v2~\cite{dinov2}, we present visualizations derived from the Principal Component Analysis (PCA) conducted on patch features extracted using our model~\name. As depicted in Figure~\ref{fig:pca_visualization}, a comparative analysis is conducted between \name~and DINO v2, both utilizing the ViT-L/14 architecture. The results demonstrate that \name~effectively accentuates the features of cars and planes, while efficiently minimizing background clutter. 

\noindent\textbf{Scaling behavior.} 
We train ViT-BViT-L models using random subsets of varying sizes: 1M, 3M, 10M, 40M, and the comprehensive 150M (measured in the number of captions). These models are trained over a reduced schedule of 300,000 steps and utilizes a smaller batch size of 2048. The outcomes of linear probing are illustrated in Figures~\ref{fig:data_scale_imagenet} and \ref{fig:data_scale_finegrained}. These results indicate that the ViT-B model delivers robust performance at the 10M scale, with diminishing returns observed beyond this point. In contrast, the ViT-L model exhibits a greater demand for data (i.e., it underperforms ViT-B at the 3M scale) and scales better with data.

\section{Discussions and Conclusion}

\textbf{Why learn from generative models?} One compelling reason is that \emph{a generative model can act like hundreds of datasets simultaneously}. Traditionally, researchers have to spend separate effort collecting datasets for different image categories, \eg, cars, flowers, cats, dogs, and so on. 
DINO v2~\cite{dinov2} achieves robust representations by curating and amalgamating numerous such datasets.
Such a process introduces complexities such as clustering and search challenges. 
In contrast, advanced text-to-image generative models like Stable Diffusion~\cite{ldm} or Imagen~\cite{imagen} have the capability to generate \textit{many} diverse datasets. 
These models provide the flexibility to produce an infinite number of samples (albeit finite diversity) and control the generation process through textual input. Thus, generative models offer a convenient and effective method for \emph{curating} training data. In our study, we harness this advantage to synthesize images encompassing a broad spectrum of visual concepts.

\noindent\textbf{What can be further improved?} 
Enhanced caption sets can be achieved through various methods, such as enriching the set of in-context examples, optimizing the sampling ratios among different concepts, and utilizing more advanced LLMs. In terms of the learning process, one approach is to distill knowledge from a larger model, and incorporate an additional high-resolution training phase (as discussed in~\cite{dinov2}) or an intermediate IN-21k fine-tuning stage (as per~\cite{beit, beitv2}). Regarding architectural improvements, the integration of SwiGLU and LayerScale, coupled with superior model initialization strategies (referenced in~\cite{eva02}), can be beneficial. However, due to limited resources and the scope of this paper not being focused on achieving the highest possible metrics, we propose these areas for further exploration in future research endeavors.

In summary, this paper studies a new paradigm for visual representation learning -- \emph{learning from generative models}. Without using any real data, ~\name~learns visual representations that are comparable with those achieved by state of the art general-purpose visual representation learners.

{
    \small
    \bibliographystyle{ieeenat_fullname}
    \bibliography{main}
}

\appendix
\clearpage
\section{Concept Sampling}

The concepts used to synthesize captions are randomly sampled from the names of various datasets. The rough ratios are presented in Table~\ref{tab:concept_sampling}. It is likely that different combinations of these ratios lead to different results, but we do not optimize over this dimension. For example, we simply concatenate IN-21k concepts with the classes of \emph{other} datasets (\eg, Caltech-101, Pets), and do uniform sampling from the concatenated list. This may lead to under-sampling for \emph{other} datasets, as the list is dominated by IN-21 classes.

\begin{table}[h]
\centering
\setlength{\extrarowheight}{1pt}
\begin{tabular}{l|c}
source & prob. \\
\shline
IN-1k              & 0.47  \\
Aircraft           & 0.05  \\
Cars               & 0.05  \\
Food               & 0.05  \\
Flowers            & 0.03  \\
Places-365, SUN397 & 0.09  \\
IN-21k and others  & 0.26 \\
\end{tabular}
\vspace{0.7em}
\caption{\textbf{Rough concept sampling probabilities.}}
\label{tab:concept_sampling}
\end{table}

\section{Implementation Details}
\subsection{Pre-training}

The setting for our final long schedule training in Section 4.2 is summarized in Table~\ref{tab:synclr_pretraining}, where models are trained for 500k steps with a batch size of 8192 captions. For ablation study present in Section 4.1, we only train for 85k steps with a batch size of 2048 captions; for the scaling plots in Section 4.3, we train all models for 300k steps with a batch size of 2048.

\begin{table}[h]
\centering
\small
\setlength{\extrarowheight}{1pt}
\begin{tabular}{l|c}
config & value \\
\shline
batch size & 8192 \\
optimizer & AdamW~\cite{adamw}  \\
peak learning rate & 2e-3 (B), 1.5e-3 (L)  \\
weight decay & 0.04 --> 0.2, cosine  \\
optimizer momentum & $\beta_1, \beta_2{=}0.9, 0.999$  \\
learning rate schedule & cosine decay \\
steps & 500k  \\
warmup steps & 80k  \\
stoch. depth~\cite{droppath} & 0.1 (B), 0.4 (L) \\
augmentation & Downsample~\cite{stablerep} + BYOL Aug.~\cite{byol} \\
\end{tabular}
\vspace{1em}
\caption{\textbf{\name~pre-training settings.}}
\label{tab:synclr_pretraining}
\end{table}

\subsection{ImageNet linear probing}

We use the \texttt{cls} token from the final transformer block as the image representation. This is different from DINO v2, which tries to concatenate \texttt{cls} token with average pooled patch tokens and sweep over whether to use multiple layers.

We follow prior work~\cite{mocov3, dino} to train the linear classifier. It has been generally observed that regularization such as weight decay hurts the performance~\cite{cmc,moco}. Therefore, we set weight decay as 0, and we sweep the $base\_lr$ over $\{0.1, 0.2, 0.5, 1, 2, 5, 10, 20, 50\}\times 10^{-2}$.

\begin{table}[h]
\centering
\small
\begin{tabular}{l|c}
config & value \\
\shline
batch size & 1024 \\
optimizer & SGD  \\
base learning rate & sweep \\
peak learning rate & $blr\times bsz/$256 \\
weight decay & 0  \\
optimizer momentum & 0.9 \\
learning rate schedule & cosine decay \\
epochs & 90  \\
augmentation & RandomResizedCrop, Flip \\
\end{tabular}
\vspace{1.0em}
\caption{\textbf{ImageNet linear probing settings.}}
\label{tab:imagenet_linear_param}
\end{table}

\subsection{End-to-End ImageNet fine-tuning}
Following common practice~\cite{beit,mae}, we append a linear classifier on top of the \texttt{CLS} token of the last transformer block, and fine-tune the whole network. We use layer-wise \emph{lr} decay~\cite{lrdecay}. Table~\ref{tab:imagenet_finetune_param} shows the settings. 

\begin{table}[h]
\centering
\small
\begin{tabular}{l|c}
config & value \\
\shline
optimizer & AdamW~\cite{adamw} \\
base learning rate & 5e-5 \\
peak learning rate & $blr\times bsz/$256 \\
optimizer momentum & $\beta_1, \beta_2{=}0.9, 0.999$ \\
layer-wise lr decay & 0.65 (B), 0.8 (L) \\
batch size & 1024 \\
learning rate schedule & cosine decay \\
warmup epochs & 20 (B), 5 (L) \\
epochs & 100 (B), 50 (L)  \\
RandAugment~\cite{Randaugment} & 9/0.5 \\
label smoothing & 0.1 (B), 0.2 (L) \\
erasing prob. & 0.25 \\
mixup~\cite{mixup} & 0.8 \\
cutmix~\cite{cutmix} & 1.0 \\
stoch. depth~\cite{droppath} & 0.1 (B), 0.3 (L) \\
test crop ratio & 0.95 (B), 1.0 (L) \\
ema & 0.9999
\end{tabular}
\vspace{1.0em}
\caption{\textbf{ImageNet end-to-end fine-tuning settings.}}
\label{tab:imagenet_finetune_param}
\end{table}

\subsection{Semantic segmentation on ADE20k}

We conduct the experiments on ADE20k~\cite{ade20k}. Following~\cite{beit,mae}, we use UperNet~\cite{upernet} as the task adaptation layer. We use the common \emph{single-scale}~\cite{beit} setup, with a resolution of 512$\times$512 for models with a patch size of 16$\times$16 and a resolution of 518$\times$518 for models with a patch size of 14$\times$14. The hyper-parameters are summarized in Table~\ref{tab:ade20k_param}.

\begin{table}[h]
\centering
\small
\begin{tabular}{l|c}
config & value \\
\shline
batch size & 32 (B), 16 (L) \\
optimizer & AdamW~\cite{adamw} \\
peak learning rate & 8e-5 \\
optimizer momentum & $\beta_1, \beta_2{=}0.9, 0.999$ \\
weight decay & 0.05 \\
layer-wise lr decay & 0.6 (B), 0.8 (L) \\
steps & 60k (B), 160k (L) \\
warmup steps & 1500 \\
stoch. depth & 0.1 (B), 0.2 (L) \\
\end{tabular}
\caption{\textbf{ADE20k semantic segmentation settings.}}
\label{tab:ade20k_param}
\end{table}

\subsection{Fine-grained linear classification}

Following prior works~\cite{simclr,byol}, we train a regularized multi-nomial logistic regression model upon the output \texttt{CLS} token. In training and testing, we do not perform any data augmentation; images are resized to 224 pixels along the shorter side, followed by a center crop of 224$\times$224. We minimize the cross-entropy objective using L-BFGS with $\ell_2$-regularization. We select this $\ell_2$-regularization constant on the validation set over 45 logarithmically spaced values between $10^{-6}$ and $10^{5}$. The maximum number of L-BFGS iterations is set to $1000$, similar as that in DINO v2~\cite{dinov2}.

\section{In-context Learning Examples}

All of the three types of in-context examples are summarized in Table~\ref{tab:example_1}, Table~\ref{tab:example_2}, and Table~\ref{tab:example_3}, respectively.

\onecolumn %
\begin{longtable}[c]{p{0.06\textwidth}@{\hspace{-1.0em}}|p{0.17\textwidth}@{\hspace{0.2em}} p{0.05\textwidth}@{\hspace{-0.9em}} p{0.7\textwidth} }
\caption{Detailed in-context learning examples for Template 1: \emph{\textcolor{darkred}{$c$} --> Caption}. Here \emph{\textcolor{darkred}{$c$}} is the concept. }
\label{tab:example_1}
\\
\toprule[1.2pt]
{\hspace{0.3em}} 1 & \textcolor{darkred}{coucal} & --> & A vibrant \textcolor{darkred}{coucal} is perched on the branch of a lush green tree, surrounded by wildflowers. \\
{\hspace{0.3em}} 2 & \textcolor{darkred}{bee eater} & --> & A lively \textcolor{darkred}{bee eater} is elegantly perched on a branch, peering intently. \\
{\hspace{0.3em}} 3 & \textcolor{darkred}{three-toed sloth} & --> & A \textcolor{darkred}{three-toed sloth} is lazily hanging from a sturdy, tropical rainforest tree. \\
{\hspace{0.3em}} 4 & \textcolor{darkred}{hay} & --> & In the serene countryside, hundreds of neatly stacked \textcolor{darkred}{hay} bales lay scattered under the softly glowing golden sunset sky. \\
{\hspace{0.3em}} 5 & \textcolor{darkred}{station wagon} & --> & A shiny, red \textcolor{darkred}{station wagon} is parked under the dappled shade of a large oak tree, highlighting its spacious and family-friendly design. \\
{\hspace{0.3em}} 6 & \textcolor{darkred}{zebra} & --> & A \textcolor{darkred}{zebra} is gallantly trotting across the vast, sunlit plains of the African savannah, creating a captivating black and white spectacle. \\
{\hspace{0.3em}} 7 & \textcolor{darkred}{vase} & --> & In the well-lit living room, a beautifully designed, delicate \textcolor{darkred}{vase} stands out as the centerpiece, exuding an aura of elegance. \\
{\hspace{0.3em}} 8 & \textcolor{darkred}{barber chair} & --> & A shiny black \textcolor{darkred}{barber chair} sits invitingly in a bustling, well-lit barbershop. \\
{\hspace{0.3em}} 9 & \textcolor{darkred}{carbonara} & --> & A heaping plate of creamy \textcolor{darkred}{carbonara} pasta topped with fresh parsley sprigs. \\
{\hspace{-0.1em}} 10 & \textcolor{darkred}{mink} & --> & In the midst of a dense forest with shimmering green leaves, a sleek \textcolor{darkred}{mink} gracefully navigates the underbrush, showcasing its rich, brown fur. \\
{\hspace{-0.1em}} 11 & \textcolor{darkred}{small white butterfly} & --> & A \textcolor{darkred}{small white butterfly} gracefully flutters amongst vibrant, blooming summer flowers. \\
{\hspace{-0.1em}} 12 & \textcolor{darkred}{christmas stocking} & --> & A vibrant red \textcolor{darkred}{Christmas stocking} is hanging delicately from a festively decorated mantelpiece. \\
{\hspace{-0.1em}} 13 & \textcolor{darkred}{horse-drawn vehicle} & --> & An antique \textcolor{darkred}{horse-drawn vehicle} is stationed amidst a peaceful country landscape, its rustic wooden structure gleaming under the warm afternoon sun. \\
{\hspace{-0.1em}} 14 & \textcolor{darkred}{ruler measuring stick} & --> & A manual craftsman is precisely measuring a wooden log with a \textcolor{darkred}{ruler stick}. \\
{\hspace{-0.1em}} 15 & \textcolor{darkred}{picket fence} & --> & A tranquil suburban scene featuring multiple white \textcolor{darkred}{picket fence}s surrounding well-maintained green lawns, punctuated by diverse, colorful flowerbeds. \\
{\hspace{-0.1em}} 16 & \textcolor{darkred}{suspension bridge} & --> & Depicting a long \textcolor{darkred}{suspension bridge}, its steel cables elegantly stretching towards the sky, connecting two ends over a scenic river. \\
{\hspace{-0.1em}} 17 & \textcolor{darkred}{brain coral} & --> & A vibrant \textcolor{darkred}{brain coral} stands out amidst the serene backdrop of underwater marine life. \\
{\hspace{-0.1em}} 18 & \textcolor{darkred}{revolver} & --> & Multiple antique \textcolor{darkred}{revolver}s lie on a wooden table, gleaming under soft, ambient light. \\
{\hspace{-0.1em}} 19 & \textcolor{darkred}{slip-on shoe} & --> & A pair of \textcolor{darkred}{slip-on shoe}s, with their sleek, black leather exterior and comfortable, cushioned interior, are neatly placed on a wooden floor. \\
{\hspace{-0.1em}} 20 & \textcolor{darkred}{hand-held computer} & --> & A \textcolor{darkred}{hand-held computer}, compact and portable, rests on a well-lit desk, surrounded by various technological paraphernalia and a steaming cup of coffee. \\
{\hspace{-0.1em}} 21 & \textcolor{darkred}{mattress} & --> & A teddy bear lying face down on a bedspread covered \textcolor{darkred}{mattress} in front of a window. \\
{\hspace{-0.1em}} 22 & \textcolor{darkred}{refrigerator} & --> & A nicely decorated kitchen with metallic \textcolor{darkred}{refrigerator} and blue counter. \\
{\hspace{-0.1em}} 23 & \textcolor{darkred}{ball} & --> & Silver \textcolor{darkred}{ball}s are lined up in the sand as people mill about in the background. \\
{\hspace{-0.1em}} 24 & \textcolor{darkred}{wheel} & --> & The motorcycle's gleaming steering \textcolor{darkred}{wheel}, vivid red door reflected in the side mirror, and a youth passing by, creating a dynamic urban tableau. \\
{\hspace{-0.1em}} 25 & \textcolor{darkred}{plane} & --> & A group of trick \textcolor{darkred}{plane}s turned upside down leaving smoke trails. \\
{\hspace{-0.1em}} 26 & \textcolor{darkred}{vehicle} & --> & Army \textcolor{darkred}{vehicle}s, including a U.S. Army jeep and aircraft in a hangar or on display \\
{\hspace{-0.1em}} 27 & \textcolor{darkred}{boy} & --> & a little \textcolor{darkred}{boy} wearing sunglasses laying on a shelf in a basement. \\
{\hspace{-0.1em}} 28 & \textcolor{darkred}{fence} & --> & a man standing near a \textcolor{darkred}{fence} as reflected in a side-view mirror of a red car. \\
{\hspace{-0.1em}} 29 & \textcolor{darkred}{wood table} & --> & A footed glass with water in front of a glass with ice tea, and green serpentine bottle with pink flowers, all on a \textcolor{darkred}{wood table} in front of chair, with a window to city view. \\
{\hspace{-0.1em}} 30 & \textcolor{darkred}{toilet} & --> & A black and white \textcolor{darkred}{toilet} sitting in a bathroom next to a plant filled with waste. \\
{\hspace{-0.1em}} 31 & \textcolor{darkred}{table lamp} & --> & A textured brass \textcolor{darkred}{table lamp}, casting a warm, golden glow, accents a cozy reading nook beside a leather armchair and a stack of books. \\
{\hspace{-0.1em}} 32 & \textcolor{darkred}{hair dryer} & --> & A modern sleek and white \textcolor{darkred}{hair dryer}, with a textured grip, stands next to a set of hairbrushes. \\
{\hspace{-0.1em}} 33 & \textcolor{darkred}{street sign} & --> & The \textcolor{darkred}{street sign}s indicate which way a car can and cannot turn while the signal light controls traffic. \\
{\hspace{-0.1em}} 34 & \textcolor{darkred}{instrument} & --> & Man dressed in Native American clothes protecting musical \textcolor{darkred}{instrument}s from the rain with an umbrella. \\
{\hspace{-0.1em}} 35 & \textcolor{darkred}{train} & --> & A man and a cow's faces are near each other as a \textcolor{darkred}{train} passes by on a bridge. \\
{\hspace{-0.1em}} 36 & \textcolor{darkred}{giraffe} & --> & A couple of large \textcolor{darkred}{giraffe} standing next to each other. \\
{\hspace{-0.1em}} 37 & \textcolor{darkred}{red admiral butterfly} & --> & a \textcolor{darkred}{red admiral butterfly}, alights upon a dew-kissed sunflower, wings glistening under the soft morning light. \\
{\hspace{-0.1em}} 38 & \textcolor{darkred}{stupa} & --> & Surrounded by verdant foliage, a white \textcolor{darkred}{stupa} rises, adorned with golden accents and intricate patterns, while devotees circle its base offering prayers. \\
{\hspace{-0.1em}} 39 & \textcolor{darkred}{elephant} & --> & A group of \textcolor{darkred}{elephant}s being led into the water. \\
{\hspace{-0.1em}} 40 & \textcolor{darkred}{bottle} & --> & Motorcycles parked on a street with a \textcolor{darkred}{bottle} sitting on the seat of the nearest the camera. \\
{\hspace{-0.1em}} 41 & \textcolor{darkred}{trombone} & --> & On a polished wooden stage, a gleaming brass \textcolor{darkred}{trombone} rests, its slide extended, next to scattered sheet music and a muted trumpet. \\
{\hspace{-0.1em}} 42 & \textcolor{darkred}{keyboard} & --> & Sleek black \textcolor{darkred}{keyboard} with illuminated backlit keys, a soft wrist rest, and a nearby wireless mouse on a textured matte desk surface. \\
{\hspace{-0.1em}} 43 & \textcolor{darkred}{bear} & --> & The brown \textcolor{darkred}{bear} sits watching another \textcolor{darkred}{bear} climb the rocks \\
{\hspace{-0.1em}} 44 & \textcolor{darkred}{snowboard} & --> & A man standing next to his \textcolor{darkred}{snowboard} posing for the camera. \\
{\hspace{-0.1em}} 45 & \textcolor{darkred}{railway} & --> & a woman and her son walking along the tracks of a disused \textcolor{darkred}{railway}. \\
{\hspace{-0.1em}} 46 & \textcolor{darkred}{sand} & --> & the waves and the \textcolor{darkred}{sand} on the beach close up \\
{\hspace{-0.1em}} 47 & \textcolor{darkred}{pixel} & --> & very colorful series of squares or \textcolor{darkred}{pixel}s in all the colors of the spectrum , from light to dark \\
{\hspace{-0.1em}} 48 & \textcolor{darkred}{cigar} & --> & a burning \textcolor{darkred}{cigar} in a glass ashtray with a blurred background. \\
{\hspace{-0.1em}} 49 & \textcolor{darkred}{music} & --> & happy girl listening \textcolor{darkred}{music} on headphones and using tablet in the outdoor cafe. \\
{\hspace{-0.1em}} 50 & \textcolor{darkred}{earring} & --> & this gorgeous pair of \textcolor{darkred}{earring}s were featured in april issue. \\
{\hspace{-0.1em}} 51 & \textcolor{darkred}{cliff} & --> & Steep \textcolor{darkred}{cliff}, jagged edges against azure sky, with seabirds soaring and waves crashing below. \\
{\hspace{-0.1em}} 52 & \textcolor{darkred}{corn cob} & --> & Fresh \textcolor{darkred}{corn cob}, golden kernels glistening with dew, nestled amid green husks in a sunlit field. \\
{\hspace{-0.1em}} 53 & \textcolor{darkred}{archaeological excavation} & --> & In this intriguing scene, archaeologists meticulously uncover ancient relics at an \textcolor{darkred}{archaeological excavation} site filled with historical secrets and enigmas. \\
{\hspace{-0.1em}} 54 & \textcolor{darkred}{formal garden} & --> & This is an immaculately kept \textcolor{darkred}{formal garden}, with perfectly trimmed hedges, colorful, well-arranged flower beds, and classic statuary, giving a vibe of tranquil sophistication. \\
{\hspace{-0.1em}} 55 & \textcolor{darkred}{veterinarians office} & --> & The busy \textcolor{darkred}{veterinarian's office} is a hive of activity with pets awaiting treatment and care. \\
{\hspace{-0.1em}} 56 & \textcolor{darkred}{elevator} & --> & A modern, well-lit \textcolor{darkred}{elevator} interior with shiny metal walls and sleek buttons. \\
{\hspace{-0.1em}} 57 & \textcolor{darkred}{heliport} & --> & Situated in a lively area, the \textcolor{darkred}{heliport} stands out with numerous helicopters taking off and landing against the city's skyline. \\
{\hspace{-0.1em}} 58 & \textcolor{darkred}{airport terminal} & --> & In the spacious \textcolor{darkred}{airport terminal}, travelers hurriedly navigate through check-ins and security, making it a hive of constant activity. \\
{\hspace{-0.1em}} 59 & \textcolor{darkred}{car interior} & --> & \textcolor{darkred}{Inside the car}, the leather seats exude luxury, contrasted by the high-tech dashboard, creating an atmosphere of sleek comfort and convenience. \\
{\hspace{-0.1em}} 60 & \textcolor{darkred}{train interior} & --> & The \textcolor{darkred}{inside of the train} offers a spacious setting with numerous comfortable seats. \\
{\hspace{-0.1em}} 61 & \textcolor{darkred}{candy store} & --> & The sweet aroma of sugared treats fills the air in a vibrant \textcolor{darkred}{candy store}, adorned with colourful candies and cheerful customers. \\
{\hspace{-0.1em}} 62 & \textcolor{darkred}{bus station} & --> & The bustling \textcolor{darkred}{bus station} thrums with restless energy, as travelers navigate through the crowded space, awaiting their journeys amid the echoes of departing buses. \\
{\hspace{-0.1em}} 63 & \textcolor{darkred}{castle} & --> & Nestled amidst towering mountains, the majestic \textcolor{darkred}{castle} spews ancient grandeur, with its stone walls and towering turrets exuding tranquility and timeless mystique. \\
{\hspace{-0.1em}} 64 & \textcolor{darkred}{palace} & --> & The grand \textcolor{darkred}{palace} exudes regality, radiant under the sun, showcasing ornate decorations, intricate sculptures, and exquisite architectural sophistication. \\
{\hspace{-0.1em}} 65 & \textcolor{darkred}{kitchen} & --> & The heart of the home unfolds in the \textcolor{darkred}{kitchen}, characterized by stainless steel appliances, navy blue cabinets, and a patterned tile backsplash. \\
{\hspace{-0.1em}} 66 & \textcolor{darkred}{raceway} & --> & The high-speed adrenaline-filled atmosphere of the \textcolor{darkred}{raceway} is pulsing with the roars of powerful engines and excited cheering fans. \\
{\hspace{-0.1em}} 67 & \textcolor{darkred}{bakery} & --> & The warm, inviting \textcolor{darkred}{bakery} is filled with the intoxicating aroma of fresh bread, assorted pastries, and brewing coffee. \\
{\hspace{-0.1em}} 68 & \textcolor{darkred}{medina} & --> & This ancient, labyrinth-like \textcolor{darkred}{medina} exudes an air of mystique with its vibrantly decorated shops lining narrow, stone-cobbled pathways. \\
{\hspace{-0.1em}} 69 & \textcolor{darkred}{skyscraper} & --> & The city skyline is dominated by towering \textcolor{darkred}{skyscraper}s, creating a captivating blend of technology and architectural innovation. \\
{\hspace{-0.1em}} 70 & \textcolor{darkred}{supermarket} & --> & The \textcolor{darkred}{supermarket} scene is lively, filled with individuals scanning shelves, children reaching for treats, and clerks restocking fresh produce. \\
{\hspace{-0.1em}} 71 & \textcolor{darkred}{closet} & --> & The compact \textcolor{darkred}{closet}, brimming with clothes and shoes, exudes a feeling of organization. \\
{\hspace{-0.1em}} 72 & \textcolor{darkred}{assembly line} & --> & In the heart of a busy factory, an orderly \textcolor{darkred}{assembly line} hums with continuous activity, filled with workers focused on their precision tasks. \\
{\hspace{-0.1em}} 73 & \textcolor{darkred}{palace room} & --> & A man in military dress uniform stands in an ornate \textcolor{darkred}{palace room} with antique furniture and Christmas decorations. \\
{\hspace{-0.1em}} 74 & \textcolor{darkred}{barn doorway} & --> & A farmer holding an animal back while another farmer stands in a \textcolor{darkred}{barn doorway}. \\
{\hspace{-0.1em}} 75 & \textcolor{darkred}{food court} & --> & A bustling \textcolor{darkred}{food court} with a variety of culinary stalls, featuring vibrant signage, aromatic dishes, and communal seating, creates a diverse dining experience. \\
{\hspace{-0.1em}} 76 & \textcolor{darkred}{mountain} & --> & Majestic \textcolor{darkred}{mountain}s, their peaks dusted with snow, overlook a serene alpine lake where hikers and photographers gather to enjoy the breathtaking scenery. \\
{\hspace{-0.1em}} 77 & \textcolor{darkred}{squash court} & --> & Against a clear glass wall, a \textcolor{darkred}{squash court} with gleaming wooden floors, white boundary lines, and two rackets awaits players. \\
{\hspace{-0.1em}} 78 & \textcolor{darkred}{subway station} & --> & Dimly lit \textcolor{darkred}{subway station} with graffiti-covered walls, commuters waiting \\
{\hspace{-0.1em}} 79 & \textcolor{darkred}{restaurant} & --> & Cozy \textcolor{darkred}{restaurant} with wooden tables, ambient lighting, patrons chatting, and plates filled with colorful dishes, framed by exposed brick walls and hanging green plants. \\
{\hspace{-0.1em}} 80 & \textcolor{darkred}{field} & --> & there is a large heard of cows and a man standing on a \textcolor{darkred}{field}. \\
{\hspace{-0.1em}} 81 & \textcolor{darkred}{aquarium} & --> & Amidst vivid coral formations, an \textcolor{darkred}{aquarium} teems with colorful fish, shimmering under soft blue lights. \\
{\hspace{-0.1em}} 82 & \textcolor{darkred}{market} & --> & A large group of bananas on a table outside in the \textcolor{darkred}{market}. \\
{\hspace{-0.1em}} 83 & \textcolor{darkred}{park} & --> & a young boy is skating on ramps at a \textcolor{darkred}{park} \\
{\hspace{-0.1em}} 84 & \textcolor{darkred}{beach} & --> & old fishing boats \textcolor{darkred}{beach}ed on a coastal \textcolor{darkred}{beach} in countryside. \\
{\hspace{-0.1em}} 85 & \textcolor{darkred}{grass} & --> & little boy sitting on the \textcolor{darkred}{grass} with drone and remote controller. \\
{\hspace{-0.1em}} 86 & \textcolor{darkred}{woven} & --> & The \textcolor{darkred}{woven} basket's intricate pattern creates a visually captivating and tactile surface. \\
{\hspace{-0.1em}} 87 & \textcolor{darkred}{knitted} & --> & The \textcolor{darkred}{knitted} blanket envelops with cozy warmth \\
{\hspace{-0.1em}} 88 & \textcolor{darkred}{flecked} & --> & The stone surface was \textcolor{darkred}{flecked}, giving it a uniquely speckled and rough appearance. \\
{\hspace{-0.1em}} 89 & \textcolor{darkred}{bubbly} & --> & The liquid gleamed, showcasing its \textcolor{darkred}{bubbly}, effervescent texture vividly. \\
{\hspace{-0.1em}} 90 & \textcolor{darkred}{cobwebbed} & --> & The dusty corner was \textcolor{darkred}{cobwebbed}, displaying years of untouched, eerie beauty. \\
{\hspace{-0.1em}} 91 & \textcolor{darkred}{stained} & --> & A weather-worn wall manifests an intriguing pattern of \textcolor{darkred}{stained} texture. \\
{\hspace{-0.1em}} 92 & \textcolor{darkred}{scaly} & --> & The image showcases a close-up of a lizard's \textcolor{darkred}{scaly}, rough texture. \\
{\hspace{-0.1em}} 93 & \textcolor{darkred}{meshed} & --> & A patterned image depicting the intricate, tightly-knit texture of \textcolor{darkred}{meshed} fabric. \\
{\hspace{-0.1em}} 94 & \textcolor{darkred}{waffled} & --> & A fresh, golden-brown waffle displays its distinct crisply \textcolor{darkred}{waffled} texture invitingly. \\
{\hspace{-0.1em}} 95 & \textcolor{darkred}{pitted} & --> & The image portrays an intriguing terrain, characterized by a \textcolor{darkred}{pitted}, moon-like surface. \\
{\hspace{-0.1em}} 96 & \textcolor{darkred}{studded} & --> & A \textcolor{darkred}{studded} leather jacket gleams, highlighting its rough, tactile texture. \\
{\hspace{-0.1em}} 97 & \textcolor{darkred}{crystalline} & --> & The picture showcases an exquisite, \textcolor{darkred}{crystalline} texture with stunning brilliance and clarity. \\
{\hspace{-0.1em}} 98 & \textcolor{darkred}{gauzy} & --> & A delicate veil of \textcolor{darkred}{gauzy} texture enhances the ethereal, dreamy atmosphere. \\
{\hspace{-0.1em}} 99 & \textcolor{darkred}{zigzagged} & --> & The photo captures the \textcolor{darkred}{zigzagged} texture, emphasizing the rhythmic, sharp-edged patterns. \\
100 & \textcolor{darkred}{pleated} & --> & A flowing skirt delicately showcasing the intricate detail of \textcolor{darkred}{pleated} texture. \\
101 & \textcolor{darkred}{veined} & --> & A detailed image showcasing the intricate, \textcolor{darkred}{veined} texture of a leaf. \\
102 & \textcolor{darkred}{spiralled} & --> & The \textcolor{darkred}{spiralled} texture of the seashell creates a captivating, tactile pattern. \\
103 & \textcolor{darkred}{lacelike} & --> & The delicate veil features an intricate, \textcolor{darkred}{lacelike} texture, exuding elegant sophistication. \\
104 & \textcolor{darkred}{smeared} & --> & A wall coated with thick, \textcolor{darkred}{smeared} paint exudes a rough texture. \\
105 & \textcolor{darkred}{crosshatched} & --> & A worn, vintage book cover, richly \textcolor{darkred}{crosshatched}, exuding old-world charm. \\
106 & \textcolor{darkred}{particle} & --> & abstract background of a heart made up of \textcolor{darkred}{particle}s. \\
\bottomrule[1.2pt]
\end{longtable}

\begin{longtable}[c]{p{0.06\textwidth}@{\hspace{-1.0em}}|p{0.17\textwidth}@{\hspace{0.2em}} p{0.05\textwidth}@{\hspace{-0.9em}} p{0.7\textwidth} }
\caption{Detailed in-context learning examples for Template 2: \emph{\textcolor{darkred}{$c$}$, $\textcolor{darkblue}{$bg$} --> caption}. Here \emph{\textcolor{darkred}{$c$}} is the concept, and \emph{\textcolor{darkblue}{$bg$}} is the background. }
\label{tab:example_2}\\
\toprule[1.2pt]
107 & \textcolor{darkred}{stick insect}, \textcolor{darkblue}{undergrowth} & --> & A \textcolor{darkred}{stick insect}, masterfully camouflaged, clings to a fern amidst the sprawling, dense \textcolor{darkblue}{undergrowth} of a lush, tropical forest. \\
108 & \textcolor{darkred}{black swan}, \textcolor{darkblue}{public garden} & --> & In the peaceful ambiance of a lush \textcolor{darkblue}{public garden}, a majestic \textcolor{darkred}{black swan} gracefully glides across a shimmering emerald-green pond. \\
109 & \textcolor{darkred}{st. bernard}, \textcolor{darkblue}{family-photo} & --> & In the heartwarming \textcolor{darkblue}{family photo}, a gregarious \textcolor{darkred}{St. Bernard} dog is seen joyfully nestled among his adoring human companions. \\
110 & \textcolor{darkred}{measuring cup}, \textcolor{darkblue}{food prep area} & --> & In the \textcolor{darkblue}{food prep area}, multiple transparent \textcolor{darkred}{measuring cup}s are neatly organized on the marble countertop. \\
111 & \textcolor{darkred}{can opener}, \textcolor{darkblue}{hotel room} & --> & A sleek, stainless steel \textcolor{darkred}{can opener} is sitting on the glossy dark-wood kitchenette counter of a modern, well-appointed \textcolor{darkblue}{hotel room}. \\
112 & \textcolor{darkred}{small white butterfly}, \textcolor{darkblue}{pond side} & --> & A delicate, \textcolor{darkred}{small white butterfly} flutters gracefully above the tranquil \textcolor{darkblue}{pond side}, creating a serene image amidst lush greenery. \\
113 & \textcolor{darkred}{hair dryer}, \textcolor{darkblue}{theatre} & --> & A sleek, professional \textcolor{darkred}{hair dryer} is positioned center stage amidst the dramatic velvet curtains and ornate details of a bustling \textcolor{darkblue}{theatre}. \\
114 & \textcolor{darkred}{water bottle}, \textcolor{darkblue}{airport} & --> & A reusable \textcolor{darkred}{water bottle} sits on the glossy surface of a bustling \textcolor{darkblue}{airport} terminal counter, amidst a backdrop of hurried travelers and departure screens. \\
115 & \textcolor{darkred}{leonberger}, \textcolor{darkblue}{horse ranch} & --> & Several \textcolor{darkred}{Leonbergers} are joyfully romping around a bustling \textcolor{darkblue}{horse ranch}. \\
116 & \textcolor{darkred}{lighter}, \textcolor{darkblue}{motorhome} & --> & In the cozy, cluttered environment of a well-traveled \textcolor{darkblue}{motorhome}, a sleek silver \textcolor{darkred}{lighter} holds dominion on the rustic wooden table. \\
117 & \textcolor{darkred}{slug}, \textcolor{darkblue}{foliage} & --> & A solitary, glistening \textcolor{darkred}{slug} meanders slowly amidst lush, dense green \textcolor{darkblue}{foliage}, leaving a slimy trail on dewy leaves in its path. \\
118 & \textcolor{darkred}{ring binder}, \textcolor{darkblue}{education department} & --> & The \textcolor{darkred}{ring binder}, filled with important documents, sits prominently on a well-organized desk in the bustling \textcolor{darkblue}{education department}. \\
119 & \textcolor{darkred}{weimaraner}, \textcolor{darkblue}{pet store} & --> & A sleek, silver-gray \textcolor{darkred}{Weimaraner} is spotted curiously sniffing around various pet supplies in a well-stocked and vibrant \textcolor{darkblue}{pet store}. \\
120 & \textcolor{darkred}{norfolk terrier}, \textcolor{darkblue}{countryside} & --> & A lively \textcolor{darkred}{Norfolk terrier} joyfully bounds across a lush, green \textcolor{darkblue}{countryside}, its red fur contrasting vividly with the vast open surroundings. \\
121 & \textcolor{darkred}{dalmatian}, \textcolor{darkblue}{apple orchard} & --> & A lively \textcolor{darkred}{Dalmatian} is playfully darting amongst the lush rows of a bountiful \textcolor{darkblue}{apple orchard}, its spots contrasting against the ruby fruits. \\
122 & \textcolor{darkred}{television}, \textcolor{darkblue}{mountain lodge} & --> & A sleek, modern \textcolor{darkred}{television} sits prominently against the rustic, wooden walls of an inviting \textcolor{darkblue}{mountain lodge}, surrounded by pine-furnished decor. \\
123 & \textcolor{darkred}{guillotine}, \textcolor{darkblue}{horror story} & --> & In the shadowy landscape of a suspenseful \textcolor{darkblue}{horror story}, a grim, menacing \textcolor{darkred}{guillotine} looms ominously, exuding a petrifying sense of imminent dread. \\
124 & \textcolor{darkred}{hot tub}, \textcolor{darkblue}{condominium} & --> & A luxurious \textcolor{darkred}{hot tub} is nestled in the private balcony of a high-rise \textcolor{darkblue}{condominium}, boasting spectacular cityscape views. \\
125 & \textcolor{darkred}{leaf beetle}, \textcolor{darkblue}{plant nurseries} & --> & A vibrant \textcolor{darkred}{leaf beetle} is diligently navigating through a lush \textcolor{darkblue}{plant nursery}, its metallic sheen contrasting against the abundant green foliage. \\
126 & \textcolor{darkred}{carolina anole}, \textcolor{darkblue}{hiking trails} & --> & A small \textcolor{darkred}{Carolina Anole} lizard basks in the warm sunlight, gracefully draped over a gnarled tree root next to a bustling \textcolor{darkblue}{hiking trail}. \\
127 & \textcolor{darkred}{girl}, \textcolor{darkblue}{laboratory} & --> & teenage \textcolor{darkred}{girl} and boy working in a \textcolor{darkblue}{laboratory} on an experiment. \\
128 & \textcolor{darkred}{tiger}, \textcolor{darkblue}{forest} & --> & Two \textcolor{darkred}{tiger}s are running together in the \textcolor{darkblue}{forest}. \\
129 & \textcolor{darkred}{sunset}, \textcolor{darkblue}{lake} & --> & Golden \textcolor{darkred}{sunset} hues reflect on a calm \textcolor{darkblue}{lake}, silhouetting a lone canoeist against a backdrop of fiery clouds. \\
130 & \textcolor{darkred}{building}, \textcolor{darkblue}{mountain} & --> & town of skyline over roofs of historic \textcolor{darkred}{building}s with the \textcolor{darkblue}{mountain}s in the background. \\
131 & \textcolor{darkred}{block plane}, \textcolor{darkblue}{weathered wood} & --> & A \textcolor{darkred}{block plane}, its sharp blade gleaming, rests on \textcolor{darkblue}{weathered wood} \\
132 & \textcolor{darkred}{olive tree}, \textcolor{darkblue}{soil} & --> & single \textcolor{darkred}{olive tree} planted in the center of a dry and cracked \textcolor{darkblue}{soil} \\
133 & \textcolor{darkred}{hamster}, \textcolor{darkblue}{pet store} & --> & A curious \textcolor{darkred}{hamster} peers out, with \textcolor{darkblue}{pet store} shelves stacked with supplies behind. \\
134 & \textcolor{darkred}{bag}, \textcolor{darkblue}{factory} & --> & plastic \textcolor{darkred}{bag}s production line in a \textcolor{darkblue}{factory}. \\
135 & \textcolor{darkred}{restaurant}, \textcolor{darkblue}{ocean} & --> & young pretty couple dining in a romantic atmosphere at \textcolor{darkred}{restaurant} on the boat with \textcolor{darkblue}{ocean} on the background \\
136 & \textcolor{darkred}{helicopter}, \textcolor{darkblue}{burning forest} & --> & a \textcolor{darkred}{helicopter} flies over a portion of \textcolor{darkblue}{burning forest}. \\
137 & \textcolor{darkred}{pipe organ}, \textcolor{darkblue}{commemoration event} & --> & striking \textcolor{darkred}{pipe organ} dominates with its notes resonating, while a somber \textcolor{darkblue}{commemoration event} unfolds in the backdrop \\
138 & \textcolor{darkred}{rotisserie}, \textcolor{darkblue}{wedding reception} & --> & \textcolor{darkred}{Rotisserie} turning golden meats, with a bustling \textcolor{darkblue}{wedding reception}, twinkling lights, and guests mingling. \\
139 & \textcolor{darkred}{duck}, \textcolor{darkblue}{taiga} & --> & A group of \textcolor{darkred}{duck}s paddle on a tranquil pond, dense \textcolor{darkblue}{taiga} and towering conifers looming in the background. \\
140 & \textcolor{darkred}{tiger beetle}, \textcolor{darkblue}{rice fields} & --> & Amidst verdant \textcolor{darkblue}{rice fields}, a shimmering \textcolor{darkred}{tiger beetle} perches prominently on a dew-kissed blade of grass. \\
141 & \textcolor{darkred}{girl}, \textcolor{darkblue}{barn} & --> & slow motion clip of a \textcolor{darkred}{girl} walking with her horse through a \textcolor{darkblue}{barn} \\
142 & \textcolor{darkred}{headmaster}, \textcolor{darkblue}{graduation ceremony} & --> & the \textcolor{darkred}{headmaster} addresses the graduating seniors during \textcolor{darkblue}{graduation ceremonies}. \\
143 & \textcolor{darkred}{businessperson}, \textcolor{darkblue}{music festival} & --> & \textcolor{darkred}{businessperson} and guest attend \textcolor{darkblue}{music festival}. \\
144 & \textcolor{darkred}{fountain}, \textcolor{darkblue}{park} & --> & Water cascades from an ornate \textcolor{darkred}{fountain}, surrounded by autumn-hued trees in a serene \textcolor{darkblue}{park}. \\
145 & \textcolor{darkred}{speedboat}, \textcolor{darkblue}{water} & --> & A sleek \textcolor{darkred}{speedboat} glides on shimmering \textcolor{darkblue}{water}s, powered by twin high-horsepower outboard motors. \\
146 & \textcolor{darkred}{pipe}, \textcolor{darkblue}{beach} & --> & a rusty water \textcolor{darkred}{pipe} on the \textcolor{darkblue}{beach}. \\
147 & \textcolor{darkred}{pretzel}, \textcolor{darkblue}{home kitchen} & --> & Golden \textcolor{darkred}{pretzel} rests on a wooden board, with a cozy \textcolor{darkblue}{home kitchen}, pots and tiled backsplash, behind. \\
148 & \textcolor{darkred}{forklift}, \textcolor{darkblue}{paper mill} & --> & A \textcolor{darkred}{forklift} transports hefty paper rolls amidst the industrial bustling \textcolor{darkblue}{paper mill}. \\
149 & \textcolor{darkred}{lotion}, \textcolor{darkblue}{therapy center} & --> & Blue \textcolor{darkred}{lotion} bottles lined up at a thalasso \textcolor{darkblue}{therapy center} by the ocean. \\
150 & \textcolor{darkred}{guinea pig}, \textcolor{darkblue}{sand dunes} & --> & \textcolor{darkred}{Guinea pig} exploring vast golden \textcolor{darkblue}{sand dunes}, with tiny footprints trailing behind. \\
151 & \textcolor{darkred}{groom}, \textcolor{darkblue}{wedding ceremony} & --> & father of \textcolor{darkred}{groom} congratulating him after the \textcolor{darkblue}{wedding ceremony}. \\
152 & \textcolor{darkred}{fishing boat}, \textcolor{darkblue}{village} & --> & \textcolor{darkred}{fishing boat}s moored at fishing \textcolor{darkblue}{village} a suburb of capital of the state, \\
153 & \textcolor{darkred}{red fox}, \textcolor{darkblue}{yard} & --> & wild \textcolor{darkred}{red fox} sitting on a partially snow covered front \textcolor{darkblue}{yard} of a house in the suburbs of a small city \\
154 & \textcolor{darkred}{grey wolf}, \textcolor{darkblue}{woodland areas} & --> & A \textcolor{darkred}{grey wolf} prowls silently, eyes alert, through dense, misty \textcolor{darkblue}{woodland areas} with moss-covered trees. \\
155 & \textcolor{darkred}{cheetah}, \textcolor{darkblue}{edges of swamplands} & --> & A \textcolor{darkred}{cheetah} crouches, poised and watchful, at the lush \textcolor{darkblue}{edges of murky swamplands}. \\
156 & \textcolor{darkred}{wine bottle}, \textcolor{darkblue}{living room} & --> & in the \textcolor{darkblue}{living room}, a person si opening a \textcolor{darkred}{wine bottle} with corkscrew with wooden barrel \\
\bottomrule[1.2pt]
\end{longtable}

\begin{longtable}[c]{p{0.06\textwidth}@{\hspace{-1.0em}}|p{0.17\textwidth}@{\hspace{0.2em}} p{0.05\textwidth}@{\hspace{-0.9em}} p{0.7\textwidth} }
\caption{Detailed in-context learning examples for Template 3: \emph{\textcolor{darkred}{$c$}$, $\textcolor{darkgreen}{$rel$} --> caption}. Here \emph{\textcolor{darkred}{$c$}} is the concept, and \emph{\textcolor{darkgreen}{$rel$}} is the relation. }
\label{tab:example_3}\\
\toprule[1.2pt]
157 & \textcolor{darkred}{product packet / packaging}, \textcolor{darkgreen}{next to} & --> & A vibrant \textcolor{darkred}{product packet}, adorned with colorful labels and intricate designs, is neatly placed \textcolor{darkgreen}{next to} an elegant crystal glass. \\
158 & \textcolor{darkred}{croquet ball}, \textcolor{darkgreen}{behind} & --> & A vivid, red \textcolor{darkred}{croquet ball} rests serenely, hiding \textcolor{darkgreen}{behind} a worn, rustic wooden fence in a sun-kissed, lush green lawn. \\
159 & \textcolor{darkred}{bassoon}, \textcolor{darkgreen}{in front of} & --> & A beautifully crafted \textcolor{darkred}{bassoon} stands elegantly \textcolor{darkgreen}{in front of} a backdrop of velvet curtains, ready to perform at a concert. \\
160 & \textcolor{darkred}{grand piano}, \textcolor{darkgreen}{above} & --> & A gorgeous, antique chandelier is suspended \textcolor{darkgreen}{above} the glossy black \textcolor{darkred}{grand piano}, illuminating it with warm, opulent light. \\
161 & \textcolor{darkred}{bolo tie}, \textcolor{darkgreen}{behind} & --> & A beautifully crafted \textcolor{darkred}{bolo tie} is casually hung, indicating its previous use, \textcolor{darkgreen}{behind} a rustic, well-polished wooden shelf. \\
162 & \textcolor{darkred}{waffle iron}, \textcolor{darkgreen}{next to} & --> & A large, black \textcolor{darkred}{waffle iron} is placed \textcolor{darkgreen}{next to} a sparkling glass jar filled with golden maple syrup on a wooden countertop. \\
163 & \textcolor{darkred}{komodo dragon}, \textcolor{darkgreen}{below} & --> & A young child grins excitedly, peering down from a secure bridge, as a colossal \textcolor{darkred}{Komodo dragon} sprawls lazily \textcolor{darkgreen}{below} in the wildlife park. \\
164 & \textcolor{darkred}{vaulted or arched ceiling}, \textcolor{darkgreen}{besides} & --> & \textcolor{darkgreen}{Besides} the grand marble statue, glimpses of an intricate \textcolor{darkred}{vaulted or arched ceiling} add to the room’s majestic charm. \\
165 & \textcolor{darkred}{gossamer-winged butterfly}, \textcolor{darkgreen}{next to} & --> & A lovely, vibrant \textcolor{darkred}{gossamer-winged butterfly} is gently perched \textcolor{darkgreen}{next to} a dew-kissed red rose in an early morning garden. \\
166 & \textcolor{darkred}{kit fox}, \textcolor{darkgreen}{in front of} & --> & A group of small, fluffy, golden \textcolor{darkred}{kit fox}es is playfully gathered \textcolor{darkgreen}{in front of} a lush, green, towering forest backdrop. \\
167 & \textcolor{darkred}{koala}, \textcolor{darkgreen}{in} & --> & A cute, fuzzy \textcolor{darkred}{koala} is visibly relaxed, nestled contentedly \textcolor{darkgreen}{in} the crook of a towering, lush green eucalyptus tree. \\
168 & \textcolor{darkred}{centipede}, \textcolor{darkgreen}{above} & --> & A vibrant green \textcolor{darkred}{centipede} is effortlessly crawling on a tree branch, positioned distinctly \textcolor{darkgreen}{above} a patch of untouched fern leaves. \\
169 & \textcolor{darkred}{mountain bike}, \textcolor{darkgreen}{above} & --> & A \textcolor{darkred}{mountain bike} is displayed prominently \textcolor{darkgreen}{above} the rustic mantlepiece, showcasing its sleek design and intricate details. \\
170 & \textcolor{darkred}{wallaby}, \textcolor{darkgreen}{above} & --> & A fluffy, brown \textcolor{darkred}{wallaby} is leaping high, appearing as if it is effortlessly floating \textcolor{darkgreen}{above} a lush, green Australian field. \\
171 & \textcolor{darkred}{giant panda}, \textcolor{darkgreen}{on} & --> & A playful \textcolor{darkred}{giant panda} is perched \textcolor{darkgreen}{on} a sturdy tree branch, munching \textcolor{darkgreen}{on} fresh green bamboo amidst the tranquil forest ambiance. \\
172 & \textcolor{darkred}{beagle}, \textcolor{darkgreen}{on} & --> & A pack of adorable \textcolor{darkred}{beagle}s are spotted lounging \textcolor{darkgreen}{on} an expansive, sunbathed meadow with colorful wildflowers sprouting around them. \\
173 & \textcolor{darkred}{beach}, \textcolor{darkgreen}{on} & --> & A vivid sunset is \textcolor{darkgreen}{on} display over a sprawling \textcolor{darkred}{beach}, casting warm hues \textcolor{darkgreen}{on} the waves gently lapping at the sandy shore. \\
174 & \textcolor{darkred}{grey whale}, \textcolor{darkgreen}{on} & --> & A voluminous \textcolor{darkred}{grey whale} is majestically breaching, its massive body \textcolor{darkgreen}{on} display against the azure backdrop of the expansive ocean. \\
175 & \textcolor{darkred}{tractor}, \textcolor{darkgreen}{in front of} & --> & A bright red \textcolor{darkred}{tractor} is parked \textcolor{darkgreen}{in front of} a rustic, weathered barn, casting long shadows under the golden afternoon sun. \\
176 & \textcolor{darkred}{cabbage}, \textcolor{darkgreen}{besides} & --> & A vibrant image portrays a lush, green \textcolor{darkred}{cabbage}, glistening with dewdrops, nestled \textcolor{darkgreen}{besides} a rustic, wooden crate full of freshly harvested vegetables. \\
\bottomrule[1.2pt]
\end{longtable}
\twocolumn

\end{document}